\documentclass[sigconf,nonacm]{acmart}
\AtBeginDocument{%
  }

\setcopyright{acmlicensed}
\copyrightyear{2026}
\acmYear{2026}
\acmDOI{XXXXXXX.XXXXXXX}
\acmISBN{978-1-4503-XXXX-X/2018/06}

\acmSubmissionID{3337}

\usepackage[table]{xcolor} % 支持 \cellcolor
\usepackage{array}         % 支持 \extrarowheight
\usepackage{booktabs}      % 支持 	oprule, \aboverulesep 等
\usepackage{multirow}
\begin{document}

\title{Euler-inspired Decoupling Neural Operator for Efficient Pansharpening}

%\author{Anonymous Author(s)}
%\email{webmaster@marysville-ohio.com}
%\affiliation{%
  %\institution{Institute for Clarity in Documentation}
 %% \city{Dublin}
 %\country{USA}
%}

%\author{Lars Th{\o}rv{\"a}ld}
%\affiliation{%
  %\institution{The Th{\o}rv{\"a}ld Group}
  %\city{Hekla}
  %\country{Iceland}}
%\email{larst@affiliation.org}

%\affiliation{%
  %\institution{Inria Paris-Rocquencourt}
  %\city{Rocquencourt}
  %\country{France}
%}

\author{Anqi Zhu}
\affiliation{%
  \institution{School of Software Technology, Zhejiang University}
  \city{Ningbo}
  \country{China}
}

\author{Mengting Ma}
\affiliation{%
  \institution{School of Software Technology, Zhejiang University}
  \city{Ningbo}
  \country{China}
}

\author{Yizhen Jiang}
\affiliation{%
  \institution{School of Software Technology, Zhejiang University}
  \city{Ningbo}
  \country{China}
}

\author{Xiangdong Li}
\affiliation{%
  \institution{School of Software Technology, Zhejiang University}
  \city{Ningbo}
  \country{China}
}

\author{Kai Zheng}
\affiliation{%
  \institution{School of Software Technology, Zhejiang University}
  \city{Ningbo}
  \country{China}
}

\author{Jiaxin Li}
\affiliation{%
  \institution{School of Computer Science and Technology , Chongqing University of Post and Telecommunications}
  \city{Chongqing}
  \country{China}
}

\author{Wei Zhang}
\affiliation{%
  \institution{School of Software Technology, Zhejiang University}
  \city{Ningbo}
  \country{China}
}

%\affiliation{%
 %\institution{Rajiv Gandhi University}
 %\state{Arunachal Pradesh}
 %\country{India}}

\renewcommand\footnotetextcopyrightpermission[1]{}
\settopmatter{printacmref=false} %remove ACM reference format
\begin{abstract}

Integrating asymmetric modalities with high-resolution structural details from panchromatic images and spectral properties from low-resolution multispectral images is a core challenge in representation learning. This is especially true in pansharpening, where the frequency difference between panchromatic (PAN) and low-resolution multispectral (LR-MS) signals often causes a severe coupling conflict. Standard operators in Cartesian coordinates struggle to harmonize these disparate features because the combined update of real and imaginary parts leads to phase drift and spectral blurring. To resolve this contradiction, we propose the Eulerian Neural Operator. This physics-inspired framework redefines asymmetric feature fusion as a continuous functional mapping on polar manifolds. Departing from standard convolutions, our approach uses Euler’s formula to decouple feature representations into magnitude and phase components. Specifically, we develop an Eulerian interaction mechanism that splits the fusion process into two specialized paths. In the first path, explicit phase rotation ensures geometric consistency through linear evolution in the circular domain. In the second path, implicit magnitude mapping calibrates spectral distributions via neural manifold approximation. Operating in the frequency domain allows our model to capture global receptive fields and exhibit a resolution-invariant property, enabling zero-shot generalization across varying spatial scales. Extensive experiments demonstrate that this lightweight architecture achieves a superior balance between physical interpretability, computational efficiency, and reconstruction quality.

\end{abstract}



 \maketitle

\section{Introduction}
Pansharpening is a fundamental task in remote sensing image processing. Conceptually, due to physical constraints, satellite sensors cannot simultaneously capture images with both high spatial resolution and rich spectral information~\cite{zhang2022progress,li2023x}. Instead, they provide image pairs consisting of a low-resolution multispectral (LR-MS) image ("colorful but blurry") and a high-resolution panchromatic (PAN) image ("black-and-white but sharp") ~\cite{ghassemian2016review,thomas2008synthesis}. The goal of pansharpening is to fuse these two images to synthesize a high-resolution multispectral (HR-MS) image that is both colorful and sharp.

\begin{figure*}[t]
    \centering
    \includegraphics[width=\linewidth]{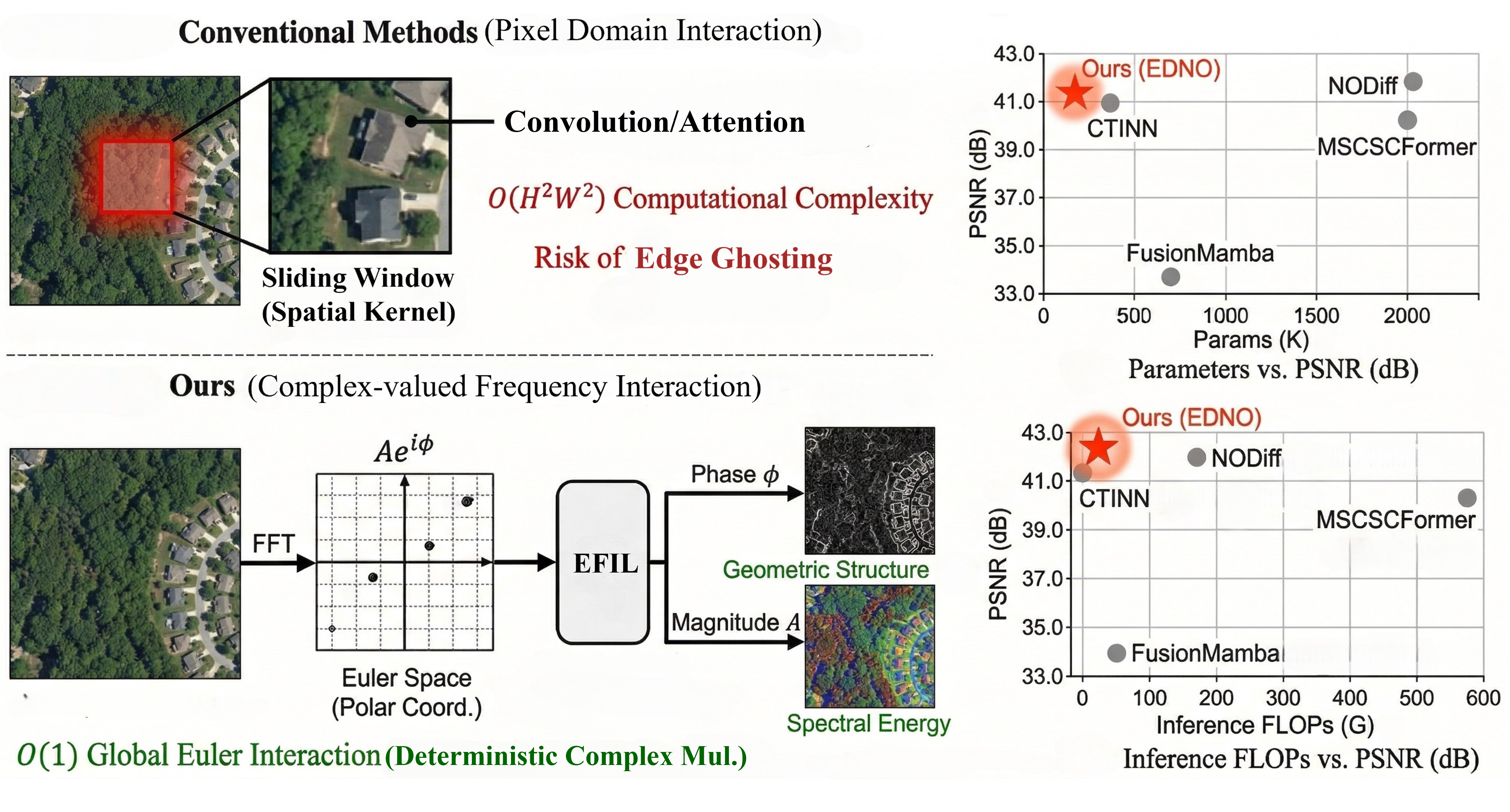}
    \caption{The motivation and performance of EDNO. (Top left) Conventional pixel-domain interactions are computationally expensive and prone to edge ghosting. (Bottom left) Our method utilizes Fast Fourier Transform (FFT) to interact in Euler space, enabling efficient global modeling of geometric structures and spectral energy. (Right) Performance benchmarking shows that EDNO (red star) dominates the upper-left corner of both PSNR-vs-Params and PSNR-vs-FLOPs charts, demonstrating superior efficiency and reconstruction quality.}
    \label{fig:teaser}
\end{figure*}

In recent years, deep learning, particularly Convolutional Neural Networks (CNNs), has significantly advanced this domain~\cite{deng2022machine,vivone2024deep,wu2021dynamic}. However, two challenges remain unaddressed. 
First is the "trade-off" dilemma caused by spectral-spatial entanglement~\cite{zhang2025s2wmamba}. Since spatial details and spectral information are deeply coupled in the pixel domain, traditional CNNs and Transformer-based architectures~\cite{Bandara2022HyperTransformer,he2021cnn,meng2022vision} often struggle to optimize both modalities simultaneously. Specifically, while Transformers excel at long-range dependency modeling, their aggressive spatial attention mechanisms often prioritize structural reconstruction at the expense of spectral integrity, leading to severe color distortion or spatial blurring~\cite{zheng2023deep,zhang2025rethinking}.
Second is the "heavy computation" bottleneck. In cutting-edge paradigms such as Transformer-based and Diffusion-based models~\cite{hou2026nodiff,cao2024diffusion,meng2023pandiff}, the pursuit of global receptive fields in the spatial domain leads to prohibitive quadratic complexity or iterative latency. These limitations stem from the rigid grid-based processing paradigm, which fails to leverage the underlying functional continuity of remote sensing scenes and the intrinsic frequency-domain sparsity of image structures. To overcome this, several studies have recently shifted toward the Neural Operators, which aim to learn a mapping between infinite-dimensional function spaces rather than discrete pixel grids. For instance, pioneering Physics-Informed Neural Operators (PINO)~\cite{liu2025PINO} attempt to bridge this gap by simulating the radiance field integration process. However, such operators encounter the dependency-complexity dilemma, i.e, they heavily rely on explicit sensor priors and incur extreme computational costs due to their intricate physical kernels, which limits their flexibility and efficiency in general remote sensing applications.

To break through the bottleneck, we shift our perspective from computationally intensive spatial-domain integral kernels to an elegant mathematical decoupling within the neural operator framework. As illustrated in Fig.~\ref{fig:teaser} (Left), while traditional pixel-domain interactions suffer from prohibitive complexity, we argue that the complex-valued frequency domain provides a more principled space for the integral kernels of a neural operator, where image structures and spectral distributions are naturally disentangled~\cite{luo2025pancomplex}.To  formulate this disentanglement, we represent frequency components in polar coordinates where phase encodes geometric shapes while magnitude determines spectral intensity. This conceptual shift leads to our proposed \textbf{Euler-inspired Decoupling Neural Operator (EDNO)}, which leverages Euler’s formula~\cite{Tian2023EulerNet,ahmad2024fpga} to map cross-modality coupling into Euler Space. Such a coordinate shift effectively linearizes the fusion task by resolving structural misalignments to deterministic phase rotations and modulating spectral variations through magnitude scaling. Consequently, EDNO achieves global Euler interaction via point-wise complex multiplication~\cite{kashefi2024novel,Guibas2021AFNO}, capturing global receptive fields with $O(1)$ parameter complexity without requiring any sensor-specific priors.

\begin{figure*}[t]
    \centering
    \includegraphics[width=\linewidth]{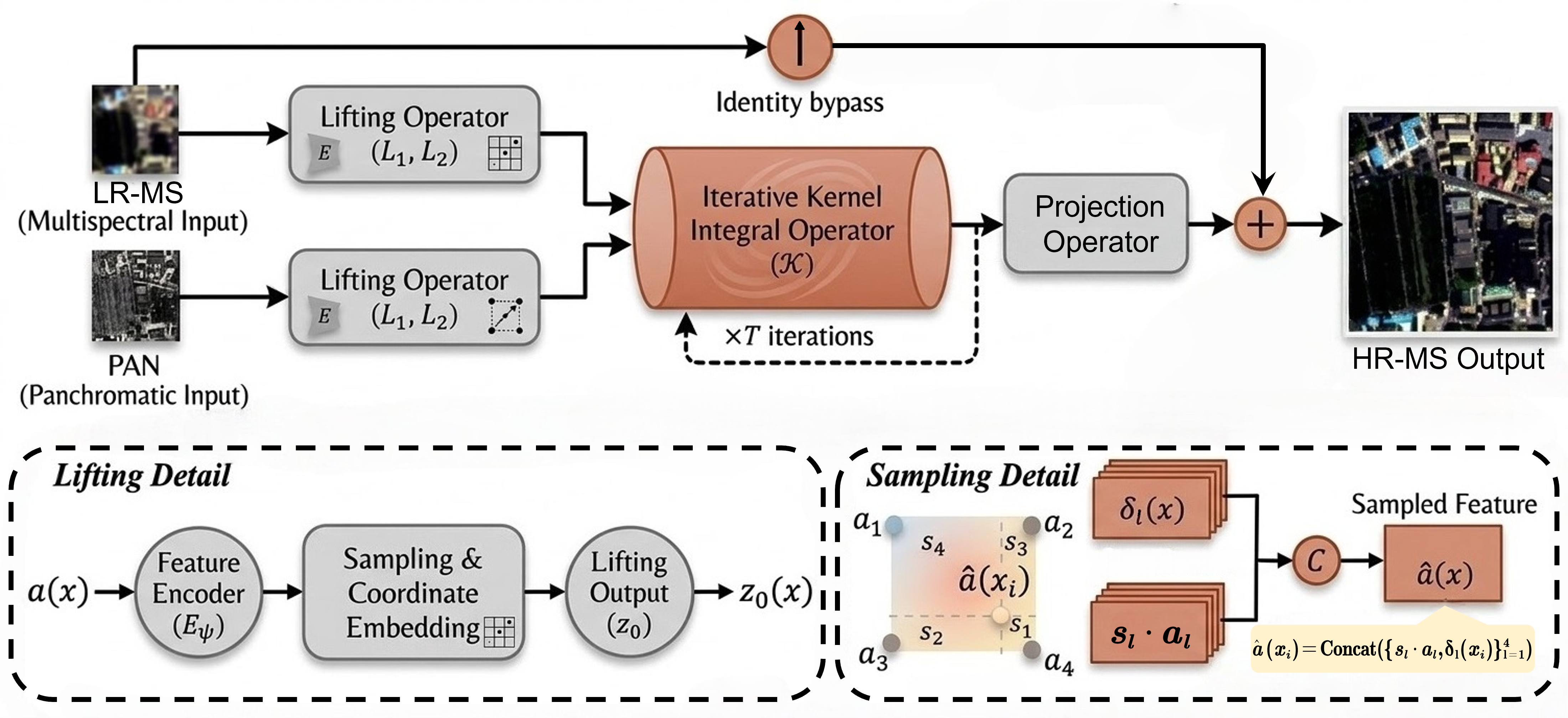}
    \caption{Overall Architecture of the proposed Euler-inspired Decoupling Neural Operator (EDNO). The EDNO framework follows a structured three-stage pipeline to achieve resolution-agnostic pansharpening. Starting with the Lifting Stage, a dual-mapping operator $L = (L_1, L_2)$ transforms the input PAN and LR-MS images into a unified, high-dimensional continuous-feature manifold $z_0(x)$ by combining a feature encoder $E_\psi$ with implicit coordinate-based sampling. Subsequently, these latent representations undergo the Iterative Kernel Integral Stage, where $T$ iterations of kernel operations ($\mathcal{K}$) progressively refine cross-modal features to resolve spectral-spatial coupling. Finally, in the Projection and Fusion Stage, the refined features are mapped back to the target output space through a projection operator and integrated with a global residual bypass to ensure maximum spectral fidelity in the reconstructed HR-MS output. The Lifting Detail and Sampling Detail (bottom) further elucidate the coordinate-aware mechanism, where neighboring grid features $a_l$ are aggregated via spatial weights $s_l$ and subsequently concatenated with fractional coordinate offsets $\delta_l$, to maintain consistency across continuous function spaces~\cite{wei2023super}.}
    \label{fig:framework}
\end{figure*}

The technical backbone of our approach is the Euler Feature Interaction Layer (EFIL), which independently refines geometric textures and spectral energy within this complex manifold. Specifically, an Explicit Feature Interaction module simulates the aforementioned phase rotations via linear weighting to align spatial features, while an Implicit Feature Interaction module employs a feed-forward network to calibrate spectral distributions. By leveraging the integral kernel properties of neural operators~\cite{Li2020FNO}, EFIL captures global receptive fields directly in the Fourier domain~\cite{kashefi2024novel,Guibas2021AFNO}, thus achieving superior parameter efficiency compared to redundant spatial convolutions~\cite{al2026fourier}. As demonstrated in Fig.~\ref{fig:teaser} (right), EDNO (indicated by the red star) establishes a superior Pareto frontier between fidelity and efficiency. It achieves a state-of-the-art PSNR exceeding $41$~dB while drastically compressing the parameter count to merely $246.9$~K. This "lightweight yet accurate" architecture allows EDNO to reconstruct sharper boundaries and more faithful colors, effectively bypassing the spatial artifacts and heavy computational overhead common in traditional black-box models.The main contributions of this paper are summarized as follows:
\begin{itemize}
\item \textbf{Novel Framework}: We propose EDNO, a lightweight and efficient frequency-domain neural operator for pansharpening, marking the first integration of Euler-inspired operator learning into this cross-modal task.

\item \textbf{Decoupled Interaction}: We develop the Euler Feature Interaction Layer (EFIL), which employs an explicit-implicit strategy to decouple the reconstruction of phase (texture) and magnitude (spectrum). This physics-aware design effectively mitigates the spectral-spatial entanglement inherent in traditional pixel-wise mapping paradigms.

\item \textbf{Superior Efficiency}: By replacing redundant spatial convolutions with frequency-domain linear combinations, our method achieves a superior efficiency-performance balance, providing a highly scalable solution for resource-constrained satellite imaging systems.
\end{itemize}

\begin{figure*}[t]
    \centering
    \includegraphics[width=\linewidth]{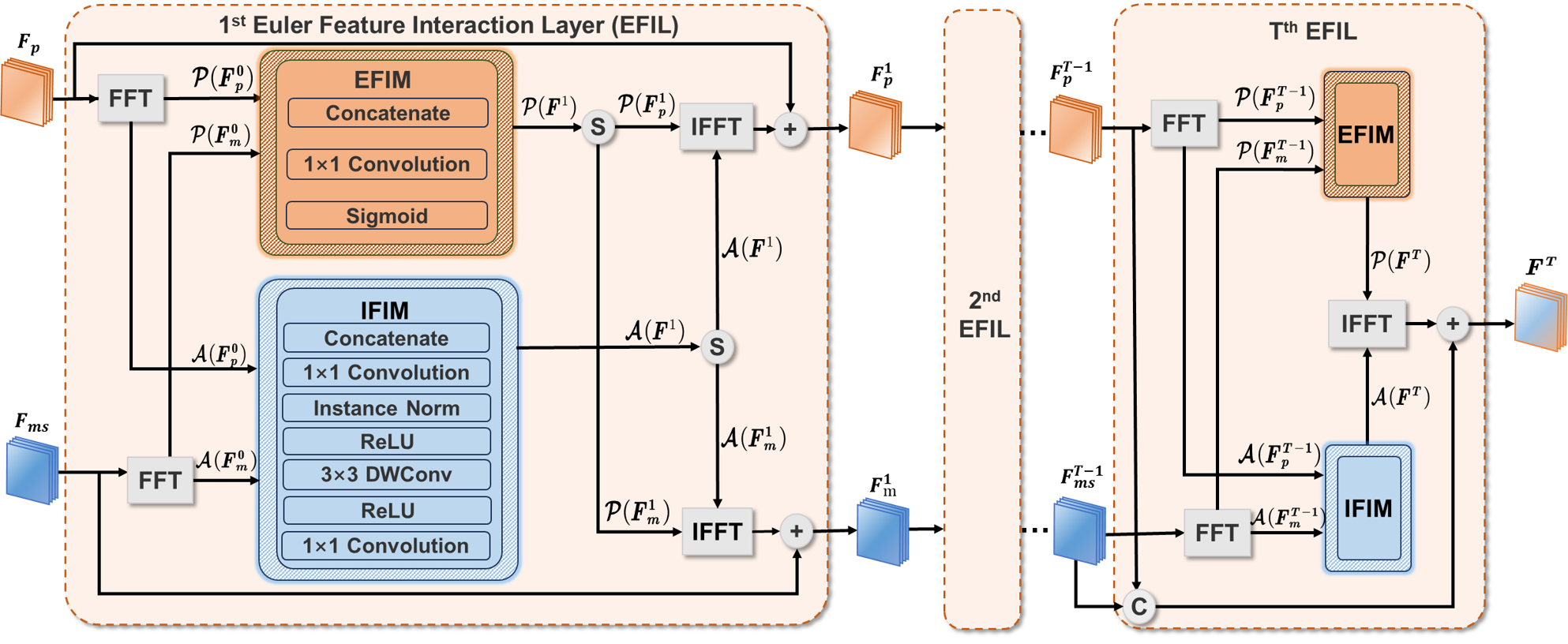}
    \caption{The detailed architecture of the Euler Feature Interaction Layer (EFIL). The input modalities are first projected onto a complex-valued frequency manifold via FFT. Within each EFIL, the Explicit Phase Interaction Module (EFIM) aligns geometric structures through phase-shifting, while the Implicit Magnitude Interaction Module (IFIM) adaptively enhances spectral energy. These decoupled components are progressively refined through $T$ iterations and finally reconstructed into the fused feature $F^T$ via IFFT.}
    \label{fig:EFIL}
\end{figure*}

\section{Related Works}

\subsection{Deep Learning for Pansharpening}

Pansharpening has transitioned from component substitution to deep learning-based paradigms~\cite{choi2010new,wang2024general,cao2024zero,li2024model,masi2016pansharpening}. Early CNN methods like PNN ~\cite{Masi2016PNN} and PanNet~\cite{Yang2017PanNet} utilized residual learning to capture spatial details. Recently, Transformer-based models such as PanFormer~\cite{Zhou2022PanFormer} and generative approaches like NODiff~\cite{hou2026nodiff} have pushed performance boundaries by leveraging long-range dependencies and iterative refinement. However, these models primarily operate in the Cartesian pixel domain, where spatial and spectral features are deeply coupled. This requires massive parameter counts~\cite{Zhou2022PanFormer} to resolve modal conflicts, leading to a high computational overhead. In contrast to these black-box pixel-mapping approaches, our EDNO introduces an Euler-inspired decoupling mechanism within the Neural Operator framework. This interaction allows for high-fidelity pansharpening while maintaining an ultra-lightweight overhead.

\subsection{Frequency-domain Fusion Strategies}
Frequency-domain learning offers global receptive fields via point-wise multiplications, providing a more efficient alternative to local convolutions~\cite{Zhou2022SFIIN,Li2023LGTEUN}. Recent works like FFPN~\cite{Chen2023FFPN} utilizes the Fourier Transform to process image components. However, these methods typically treat real and imaginary parts as independent channels in a standard CNN, a "black-box" approach that ignores the inherent physical properties of the complex spectrum. In contrast, our EDNO treats the frequency domain as a complex-valued polar space. By explicitly utilizing Euler’s formula, we decouple geometric morphology (phase) from spectral energy (magnitude), achieving a more principled and lightweight solution than existing Cartesian-based frequency methods~\cite{Hou2023BIM}.

\subsection{Neural Operators}
Neural Operators, such as the Fourier Neural Operator (FNO) ~\cite{Li2020FNO} and AFNO ~\cite{Guibas2021AFNO}, learn mappings between function spaces to ensure discretization-invariance. While NODiff ~\cite{hou2026nodiff} integrates operator learning as a denoising backbone within a diffusion loop, it still relies on costly iterative processes to synthesize textures. A very recent advancement, PINO~\cite{liu2025PINO}, incorporates sensor imaging physics into the neural operator to model continuous radiance fields. However, PINO's reliance on explicit sensor priors (e.g., spectral responsivity) and the computational intensity of its band-wise integration limit its generalizability and efficiency. Our work departs from these paradigms by leveraging the EulerNet concept ~\cite{Tian2023EulerNet} to linearize non-linear interactions into simple rotations and scaling. By bridging Euler-inspired operators with pansharpening, we achieve a "one-step" high-performance fusion without requiring sensor-specific parameters. With only $246.9$ K parameters, EDNO represents a significant efficiency leap over both physics-heavy operators like PINO.

\section{Methodology}

\subsection{Overall Framework: A Three-Stage Pipeline}

As illustrated in Fig.~\ref{fig:framework}, the proposed Euler-inspired Decoupling Neural Operator (EDNO) follows a structured three-stage pipeline consisting of lifting, iterative kernel integral, and projection operations. The process begins with the lifting operation, which maps the PAN and LR-MS images from their pixel space into a high-dimensional continuous-feature manifold. These latent features then undergo $T$ iterations of kernel integral layers, which serve as the core for cross-modal feature interaction and progressive refinement. Subsequently, the projection operation transforms these fused representations back into the target output space. To maintain spectral fidelity and ensure stable training, we incorporate a global residual connection by adding the upsampled LR-MS image to the projected features. The architectural innovation lies in the kernel integral layer, which we reformulate as the Euler feature interaction layer.

\subsection{Continuous-Space Neural Operator in Frequency Domain}
\label{sec:no}
Standard CNN-based pansharpening methods are often limited by local receptive fields, as their fixed-pixel kernels struggle to align structures across varying scales~\cite{yin2025cascaded,xia2026swift,liu2020remote}. To overcome this, we leverage the continuous-space nature of neural operators~\cite{Li2020FNO}. Unlike traditional networks that operate on discrete grids, EDNO learns to map between continuous function spaces. This ensures that the fusion process remains consistent regardless of the underlying discretization, which is a critical advantage for preserving sub-pixel structural details in remote sensing imagery.Specifically, we approximate the ideal fusion mapping using a neural network $G_{\theta}$. By shifting the computation to the frequency domain, we transform spatial convolutions into element-wise operations. This allows the model to perceive the image as a continuous spectrum rather than a collection of isolated pixels, enabling an infinite receptive field to capture long-range structural dependencies with minimal computational cost. We approximate the neural operator $G$ by training a network $G_{\theta}:A\rightarrow U$ where $A$ is defined on a bounded domain $D$ and the inputs are vector-valued functions. The iterative structure of $G$ is defined as follows:

\begin{equation}z_0\left( x \right) = L\left( x,a\left( x \right) \right) = \mathcal{C}(L_1(a(x)), L_2(x)),\end{equation}\begin{equation}z_{t+1}\left( x \right) = \sigma \left( W_tz_t\left( x \right) + (K_t(z_t; \varPhi))(x) \right),\end{equation}
\begin{equation}u\left( x \right) = P(z_T(x)),
\end{equation}
where $L$ is the lifting operator, $C$ is the concatenation operation, $W_t$ is a point-wise linear transformation, $K_t$ is the kernel integral operator, and $P$ is the projection operator. Inspired by the lifting scheme proposed in SRNO~\cite{wei2023super}, we address the inherent complexity of remote sensing images by applying this approach to lift the LR-MS input.The lifting operator $L = (L_1, L_2)$ facilitates the transition to continuous space: $L_1$ serves as a feature encoder to extract latent spectral-spatial representations from the raw pixel intensity $a(x)$, while $L_2$ acts as a coordinate-based embedding that incorporates the continuous spatial query $x$. By applying this hybrid lifting scheme, the LR-MS features are effectively projected onto a unified, high-dimensional continuous manifold $z_0(x)$.For the sampling of PAN feature maps, we employ bilinear interpolation.

However, processing the continuous manifold $z_t(x)$ directly in the spatial domain is computationally expensive for capturing global structures. To solve this, we shift the kernel integral $\mathcal{K}$ to the frequency domain, where features are naturally represented as complex numbers. This frequency view provides a simple but powerful way to separate spatial and spectral information: the phase component represents the image's geometric shapes and textures, while the magnitude component controls the color and spectral intensity. Based on this physical intuition, we propose \textbf{the Euler Feature Interaction Layer (EFIL)} in the following section. The EFIL explicitly decouples these two components to resolve the long-standing conflict between spatial detail and spectral accuracy.

\subsection{Euler Feature Interaction Layer (EFIL)}

\subsubsection{Motivation: Why Polar Decoupling in Continuous Space?}
In conventional Cartesian representations involving real and imaginary components, spatial structures and spectral intensities are coupled, making direct fusion prone to spectral bleeding or blurred textures. We argue that the complex-valued polar domain ($\mathcal{A}, \mathcal{P}$) offers a more physics-aligned manifold for continuous-space interaction. Grounded in Fourier theory, the phase ($\mathcal{P}$) captures spatial texture while the magnitude ($\mathcal{A}$) governs spectral distributions ~\cite{lin2025alphapre,zhang2024dmfourllie}. By adopting Euler’s formula~\cite{sikdar2022fully}, the pansharpening task can be decomposed into two operations, i.e., structural alignment through deterministic phase rotations and spectral calibration via magnitude scaling~\cite{Tian2023EulerNet}. Such a polar decoupling linearizes the fusion manifold, effectively bypassing the feature entanglement inherent in pixel-domain models.

\subsubsection{Overview of EFIL layer} The EFIL serves as the computational core of our framework, translating the aforementioned physical intuition into a dual-stream interaction paradigm. As illustrated in Fig.~\ref{fig:framework}, the process begins by projecting spatial features into the frequency domain via the Fast Fourier Transform (FFT) applied to the latent representations of both PAN and LR-MS. To resolve the decoupled components within the Euler space, we design two specialized modules: the Explicit Feature Interaction Module (EFIM) and the Implicit Feature Interaction Module (IFIM). Leveraging the fact that spatial displacements manifest as deterministic rotations within the phase manifold, the EFIM resolves structural misalignments through point-wise phase-shifting, while the IFIM adaptively learns the spectral mapping and magnitude scaling via a deep neural network to accommodate highly non-linear, sensor-dependent spectral energy transfers~\cite{wang2024zero}. By refining spatial details and spectral fidelity independently through this dual-path design, the EFIL offers a more generalized and efficient solution than traditional architectures that rely on stacking local spatial convolutions.

\subsubsection{Explicit Feature Interaction Module (EFIM)}
The EFIM is designed to resolve structural morphology by manipulating phase components. Based on the Fourier Shift Theorem, spatial displacement $\Delta \mathbf{x}$ is equivalent to a linear phase shift in the frequency domain. To calibrate these offsets, the EFIM extracts the phase maps $\mathcal{P}(F_p)$ and $\mathcal{P}(F_{ms})$ and learns a pixel-wise phase-shifting field. Specifically, as shown in Fig.~\ref{fig:EFIL}, we employ a $1 \times 1$ convolution to perform a weighted summation of the concatenated phases, followed by a Sigmoid activation $\sigma(\cdot)$ to normalize the fused phase distribution:
\begin{equation}
\mathcal{P}_{fused} = \sigma\left(\text{Conv}_{1\times1}\left( \text{Cat}\left[ \mathcal{P}(F_p), \mathcal{P}(F_{ms}) \right] \right) \right).
\end{equation}
By transforming spatial alignments into deterministic phase rotations, this module bypasses the translation-invariance limitations of spatial convolutions, achieving superior geometric consistency with minimal parameters.

\subsubsection{Implicit Feature Interaction Module(IFIM)} The IFIM adaptively calibrates the magnitude components $\mathcal{A}(F_p)$ and $\mathcal{A}(F_{ms})$ to ensure spectral fidelity. Given that spectral energy transfer is inherently non-linear and sensor-dependent, we treat this process as an adaptive amplitude scaling problem. To eliminate scale discrepancies between modalities, as shown in Fig.~\ref{fig:EFIL}, the IFIM first utilizes Instance Normalization (IN) on the mixed magnitude features. Subsequently, a $3 \times 3$ depth-wise convolution ($W_{dw}$) acts as a learnable spectral filter to suppress high-frequency noise while modulating the energy distribution:
\begin{equation}
\mathcal{A}_{fused} = W_{proj} \times \delta \left( W_{dw} \times \delta \left( IN\left( W_{mix} \times\text{Cat}\left[ \mathcal{A}(F_p), \mathcal{A}(F_{ms}) \right] \right) \right) \right),
\end{equation}
where $\delta$ denotes the ReLU activation. This design ensures that the model preserves the intrinsic spectral energy of LR-MS bands while precisely integrating high-frequency intensity details injected from the PAN.

\subsubsection{Iterative Refinement and Final Reconstruction}
After the decoupling and interaction in the frequency domain, the updated phase and magnitude are recombined, and the Inverse Fast Fourier Transform (IFFT) maps the features back to the spatial domain. Through $T$ iterations, the model progressively refines the cross-modal features, distilling sharp textures into the spectral backbone. For the final layer, the fused frequency representation is directly inverted to yield the high-fidelity output:
\begin{equation}
F_{out} = \mathcal{F}^{-1} \left(\mathcal{A}_{fused}, \mathcal{P}_{fused} \right).
\end{equation}
This iterative refinement, empowered by Euler decoupling, ensures that EDNO achieves state-of-the-art fidelity while remaining significantly more lightweight than Transformer-based counterparts.

\begin{table*}[!h]
\caption{Evaluation indexes of different methods on WorldView-2 (WV-2) dataset.}
\centering
\setlength{\extrarowheight}{0pt}
\addtolength{\extrarowheight}{\aboverulesep}
\addtolength{\extrarowheight}{\belowrulesep}
\setlength{\aboverulesep}{0pt}
\setlength{\belowrulesep}{0pt}
\resizebox{\linewidth}{!}{
% 1. 删除了原本开头的 c|
\begin{tabular}{c|cc|ccccc|ccc} 
\toprule
% 2. 删除了第一列对应的 \cellcolor 和 &
\rowcolor[rgb]{0.922,0.922,0.922} {\cellcolor[rgb]{0.922,0.922,0.922}}                         & {\cellcolor[rgb]{0.922,0.922,0.922}}                            & {\cellcolor[rgb]{0.922,0.922,0.922}}                          & \multicolumn{5}{c|}{Reduced-Resolution}                                                  & \multicolumn{3}{c}{Full-Resolution}                  \\
\rowcolor[rgb]{0.922,0.922,0.922} \multirow{-2}{*}{{\cellcolor[rgb]{0.922,0.922,0.922}}Method} & \multirow{-2}{*}{{\cellcolor[rgb]{0.922,0.922,0.922}}Params (K)} & \multirow{-2}{*}{{\cellcolor[rgb]{0.922,0.922,0.922}}FLOPs (G)} & PSNR$\uparrow$             & SSIM$\uparrow$             & $Q_4\uparrow$                & SAM$\downarrow$          & ERGAS$\downarrow$           & $D_\lambda\downarrow$        & $D_s\downarrow$            & QNR$\uparrow$              \\ 
\hline
% 下面每一行都删除了开头的 {\cellcolor...} & 以及 Full/Binary 的 multirow 逻辑
 SFIM ~\cite{Liu2000SFIM}                                                      & -                                                        & -                                                       & 32.6334          & 0.8728          & 0.5159          & 0.0597          & 3.1919          & 0.0737          & 0.0899          & 0.8439           \\
 Wavelet ~\cite{King2001wavelet}                                                    & -                                                          & -                                                        & 32.1992          & 0.8500          & 0.4577          & 0.0638          & 3.3799          & 0.0968          & 0.1020          & 0.8126           \\
 IHS ~\cite{Faragallah2018IHS}                                                    & -                                                          & -                                                        & 32.8250          & 0.8775          & 0.5305          & 0.0637          & 3.0585          & 0.0874          & 0.1187          & 0.8053           \\
\hline
 GPPNN ~\cite{Xu2021GPPNN}                                                      & 119.8                                                          & 11.2                                                         & 40.5086          & 0.9698          & 0.8009          & 0.0275          & 1.2060          & 0.0670          & 0.0785          & 0.8607           \\
 CTINN ~\cite{Zhou2022CTINN} & 378.2       & 13.3    & 41.2019       & 0.9735         & 0.8149          & 0.0246          & 1.0879          & 0.0658         & 0.0776          & 0.8632         \\
 PANFormer ~\cite{Zhou2022PanFormer} & 1530.3         & 12.0                                                        & 41.3495          & 0.9731         & 0.8237         & 0.0242        & 1.0621         & \textbf{0.0628}        & 0.0844         & 0.8590        \\
  MSCSCFormer ~\cite{Ye2024MSCSCformer}                                                 & 1950.7                                                        & 582.4                                                       & 40.5338          & 0.9691          & 0.7963          & 0.0270          & 1.2016          & 0.0654          & 0.0772          & 0.8634           \\ 

 FusionMamba ~\cite{Xie2024FusionMamba}                                                       & 730.0                                                        & 31.0                                                         & 33.9160          & 0.9013          & 0.6182          & 0.0558          & 2.2484          & 0.1191          & 0.0916          & 0.8014           \\
   UGCC ~\cite{Zeng2025Cross-Modal}                                                        & 233.2                                                         & 307.5                                                       & 35.8744          & 0.9296          & 0.6827          & 0.0460          & 1.9253          & 0.0887          & 0.0872          & 0.8327           \\
NODiff ~\cite{hou2026nodiff}                                                       & 2132.9    & 77.8                                                         & \textbf{42.1081}          & 0.9730          & 0.8245          & 0.0246          & \textbf{0.9970}          & 0.0644          & 0.0771          & 0.8639           \\
 	\textbf{Ours}                                                          & \textbf{264.9}                                                         & \textbf{2.4}                                                         & 	41.6488 & 	\textbf{0.9747} & 	\textbf{0.8262} & 	\textbf{0.0236} & 	1.0259 & 	0.0647 & \textbf{0.0768}          & \textbf{0.8642}           \\
\bottomrule
\end{tabular}
}
\label{tab:wv2}
\end{table*}

\subsection{Loss Function}

To achieve high-fidelity reconstruction with spatial-spectral consistency, we define a hybrid objective function $\mathcal{L}_{total}$:
\begin{equation}
\mathcal{L}_{total} = \mathcal{L}_{spa} + \lambda \mathcal{L}_{freq},
\end{equation}
where $\mathcal{L}_{spa}$ and $\mathcal{L}_{freq}$ denote spatial and frequency constraints, respectively. We empirically set $\lambda = 0.1$.

\emph{Spatial-Domain Loss.} We employ the $\ell_1$ norm to minimize the pixel-level discrepancy between the fused output $\mathbf{H}$ and the ground truth $\mathbf{G}$, promoting sharper edges and reducing artifacts:
\begin{equation}
\mathcal{L}_{spa} = \left| \mathbf{G} - \mathbf{H} \right|_1.
\end{equation}

\emph{Frequency-Domain Consistency Loss.} Although $\mathcal{L}_{spa}$ ensures intensity alignment, it often results in over-smoothed textures. Given that EDNO explicitly decouples phase and magnitude, we introduce $\mathcal{L}_{freq}$ to provide direct supervision in the frequency domain. By transforming both $\mathbf{H}$ and $\mathbf{G}$ via the Fast Fourier Transform (FFT), we compute the Mean Absolute Error (MAE):

\begin{equation}\mathcal{L}_{freq} = \left| \mathcal{F}(\mathbf{H}) - \mathcal{F}(\mathbf{G}) \right|_1.
\end{equation}
By penalizing spectral residuals, $\mathcal{L}{freq}$ constrains the network to preserve the intrinsic spectral distribution and structural morphology of remote sensing scenes.

\section{Experiments}
\subsection{Experimental settings}
\subsubsection{Datasets}
To validate our proposed approach, we conduct experiments on the publicly available dataset~\cite{li2023local}, which comprises data from the WorldView-3 (WV-3), WorldView-2 (WV-2), and GaoFen-2 (GF-2) satellites.  For each satellite sensor, the relevant low-resolution dataset is generated by downsampling according to the Wald protocol~\cite{wald1997fusion}. Our experiments are performed in both reduced-and full-resolution settings.

\subsubsection{Implementation Details}
EDNO is trained for $1050$ epochs on a single NVIDIA RTX A$5000$ GPU. For optimization, we employ the Adam optimizer with a base learning rate of $10^{-4}$ and momentum parameters set to $(\beta_1, \beta_2) = (0.9, 0.999)$. The stability constant $\varepsilon$ is set to $10^{-8}$, and the batch size is maintained at $8$ throughout the training process. \emph{Due to page limitations, experimental results on the WV-3 dataset are presented in the Supplementary material}.

\subsubsection{Evaluation Metrics}
We select five reference metrics to evaluate reduced-resolution performance, including PSNR~\cite{Alain2010PSNR}, SSIM~\cite{Zhou2004SSIM}, $Q_4$/$Q_8$~\cite{Garzelli2009Q4}, SAM~\cite{Yuhas1992SAM}, and ERGAS~\cite{Alparone2007ERGAS}. For full-scale performance evaluation, we use three non-reference metrics, including $D_\lambda$, $D_s$, and QNR~\cite{Alparone2008QNR}.

\begin{table*}[!h]
\caption{Evaluation indexes of different methods on GaoFen-2 (GF-2) dataset.}
\centering
\setlength{\extrarowheight}{0pt}
\addtolength{\extrarowheight}{\aboverulesep}
\addtolength{\extrarowheight}{\belowrulesep}
\setlength{\aboverulesep}{0pt}
\setlength{\belowrulesep}{0pt}
\resizebox{\linewidth}{!}{
% 1. 删除了原本开头的 c|
\begin{tabular}{c|cc|ccccc|ccc} 
\toprule
% 2. 删除了第一列对应的 \cellcolor 和 &
\rowcolor[rgb]{0.922,0.922,0.922} {\cellcolor[rgb]{0.922,0.922,0.922}}                         & {\cellcolor[rgb]{0.922,0.922,0.922}}                            & {\cellcolor[rgb]{0.922,0.922,0.922}}                          & \multicolumn{5}{c|}{Reduced-Resolution}                                                  & \multicolumn{3}{c}{Full-Resolution}                  \\
\rowcolor[rgb]{0.922,0.922,0.922} \multirow{-2}{*}{{\cellcolor[rgb]{0.922,0.922,0.922}}Method} & \multirow{-2}{*}{{\cellcolor[rgb]{0.922,0.922,0.922}}Params (K)} & \multirow{-2}{*}{{\cellcolor[rgb]{0.922,0.922,0.922}}FLOPs (G)} & PSNR$\uparrow$             & SSIM$\uparrow$             & $Q_4\uparrow$                & SAM$\downarrow$          & ERGAS$\downarrow$           & $D_\lambda\downarrow$        & $D_s\downarrow$            & QNR$\uparrow$              \\ 
\hline
% 下面每一行都删除了开头的 {\cellcolor...} & 以及 Full/Binary 的 multirow 逻辑
 SFIM ~\cite{Liu2000SFIM}                                                      & -                                                        & -                                                       & 34.7715          & 0.8572          & 0.4584          & 0.0657          & 4.2073          & 0.0737          & 0.0899          & 0.8439           \\
 Wavelet ~\cite{King2001wavelet}                                                    & -                                                          & -                                                        & 33.9208          & 0.8197          & 0.4033          & 0.0695          & 4.6445          & 0.0968          & 0.1020          & 0.8126           \\
 IHS ~\cite{Faragallah2018IHS}                                                    & -                                                          & -                                                        & 35.2315          & 0.8837          & 0.5217          & 0.0661          & 3.9912          & 0.0874          & 0.1187          & 0.8053           \\
\hline
 GPPNN ~\cite{Xu2021GPPNN}                                                      & 119.8                                                          & 11.2                                                         & 43.5980          & 0.9764          & 0.8663          & 0.0326          & 1.5218          & 0.0670          & 0.0785          & 0.8607           \\
  CTINN ~\cite{Zhou2022CTINN}                                                       & 378.2 & 13.3   & 44.3181          & 0.9788      & 0.8719     & 0.0291          & 1.4050      & 0.0699      &0.1561          & 0.7841      \\
 PANFormer ~\cite{Zhou2022PanFormer}                                                       & 1530.3                                                          & 12.0 & 44.8501       & 0.9805          & \textbf{0.8865}         &  0.0271        & 1.3337          & 0.0670         & 0.1806         & 0.7639        \\
  MSCSCFormer ~\cite{Ye2024MSCSCformer}                                                 & 1950.7                                                        & 582.4                                                       & 42.6871          & 0.9696          & 0.8350          & 0.0360          & 1.7149          & 0.0654          & 0.0772          & 0.8634           \\ 
 FusionMamba ~\cite{Xie2024FusionMamba}                                                       & 730.0                                                         & 31.0                                                        & 39.3506          & 0.9442          & 0.7644          & 0.0576          & 2.3823          & 0.1191          & 0.0916          & 0.8014           \\
   UGCC ~\cite{Zeng2025Cross-Modal}                                                        & 233.2                                                         & 307.5                                                       & 40.5620          & 0.9586          & 0.7803          & 0.0448          & 2.1717          & 0.0887          & 0.0872          & 0.8327           \\
  NODiff ~\cite{hou2026nodiff}                                                      & 2132.9    & 77.8  
  
  & \textbf{45.7187}    & 0.9789          & 0.8804          & 0.0279            &1.3071          & \textbf{0.0645}          & 0.0755          & \textbf{0.8711}     \\

 	\textbf{Ours}                                                          & 	\textbf{264.9}                                                          & 	\textbf{2.4}                                                         & 45.1281 & 	\textbf{0.9812} & 	0.8812 & 	\textbf{0.0270} & 	\textbf{1.2915} & 	0.0676 &   \textbf{0.0738}        & 0.8648           \\
\bottomrule
\end{tabular}
}
\label{tab:gf2}
\end{table*}

\begin{figure*}[t]
    \centering
\includegraphics[width=\linewidth]{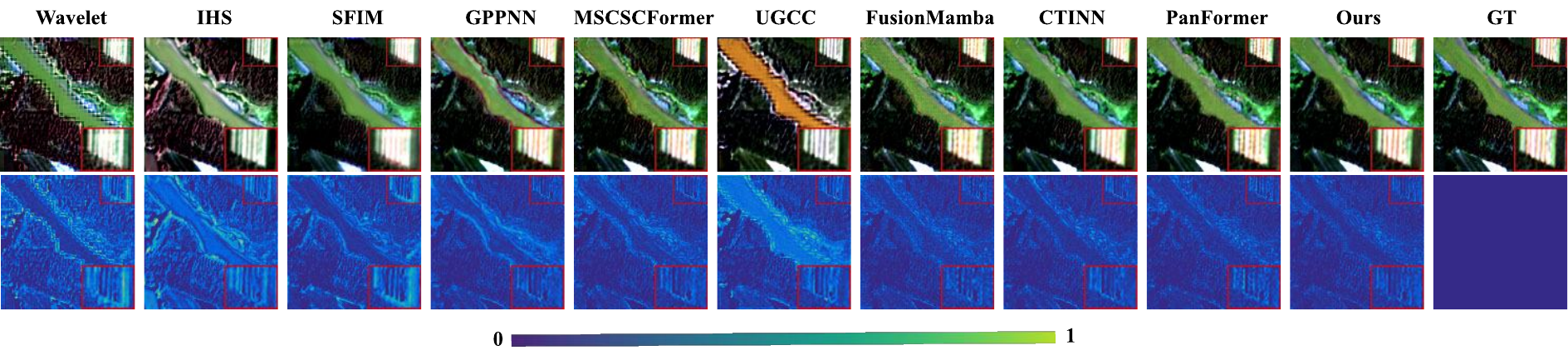}
    \caption{Visual comparison between our model and other methods on WorldView-2 (WV-2) example. The top line represent the reconstructed results, and the bottom line represents the corresponding MAE maps.}
    \label{fig:Vis-WV2-reduce}
\end{figure*}

\begin{figure*}[t]
    \centering
\includegraphics[width=\linewidth]{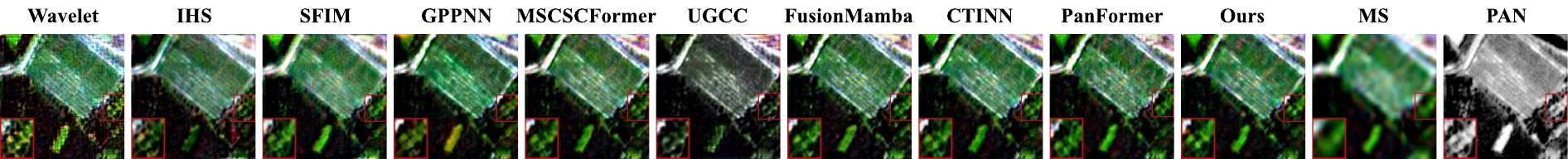}
    \caption{Visual comparison on WorldView-2 (WV-2) example at full resolution.}
    \label{fig:Vis-WV2-full}
\end{figure*}

\subsection{Comparison With SOTA Methods}
\subsubsection{Benchmark}
To evaluate the performance of the proposed method, we compare our model with $10$ state-of-the-art pansharpening methods. These encompass $3$ traditional approaches: SFIM~\cite{Liu2000SFIM}, Wavelet~\cite{King2001wavelet} and IHS~\cite{Faragallah2018IHS}; as well as $7$ deep learning-based models: GPPNN~\cite{Xu2021GPPNN}, CTINN~\cite{Zhou2022CTINN}, PANFormer~\cite{Zhou2022PanFormer}, MSCSCFormer~\cite{Ye2024MSCSCformer}, FusionMamba~\cite{Xie2024FusionMamba}, UGCC~\cite{Zeng2025Cross-Modal} and NODiff~\cite{hou2026nodiff}. Note that while Physics-Informed Neural Operator (PINO)~\cite{liu2025PINO} is a related work, it is excluded from our quantitative comparison primarily because it requires explicit sensor-specific priors, such as spectral responsivity $R_b(\lambda)$, to model the radiance field. Since our EDNO is designed as a sensor-agnostic, data-driven framework that does not rely on such internal physical parameters, a direct comparison is precluded. Furthermore, the lack of an official open-source implementation for PINO prevents a fair and reproducible benchmark.

\subsubsection{Quantitative Comparison}
Tab.~\ref{tab:wv2} and Tab.~\ref{tab:gf2} report the quantitative results across the WorldView-2 and GaoFen-2 datasets, highlighting the superior efficiency and fidelity of EDNO. Compared to state-of-the-art models, our method achieves a remarkable balance between reconstruction accuracy and computational cost. Specifically, on the WorldView-2 dataset, EDNO achieves a PSNR of $41.6488$~dB, which is significantly higher than traditional CNN-based models like GPPNN ($40.5086$~dB) and even outperforms complex architectures such as MSCSCFormer ($40.5338$~dB). While the diffusion-based model NODiff yields the highest PSNR ($42.1081$~dB), it requires a prohibitive $2132.9$~K parameters and $77.8$~G FLOPs. Similar trends are observed on the GaoFen-2 dataset, where EDNO outperforms heavy-weight competitors like FusionMamba and UGCC in terms of both spectral and spatial consistency. Most impressively, EDNO's FLOPs ($2.4$~G) remain the lowest among all deep learning-based methods, being only $1/5$ of GPPNN and $1/40$ of NODiff. These results systematically demonstrate that our Euler-inspired decoupling paradigm allows for an exceptional Pareto trade-off, delivering high-fidelity pansharpening results with an ultra-lightweight architectural overhead.

\subsubsection{Visual Comparison}
Fig.\ref{fig:Vis-WV2-reduce} and Fig.\ref{fig:Vis-WV2-full} illustrate the qualitative results of EDNO against state-of-the-art methods on WorldView-2. In the reduced-resolution evaluation (Fig.\ref{fig:Vis-WV2-reduce}), the Mean Absolute Error (MAE) maps reveal that conventional approaches and even deep models like GPPNN or PanFormer exhibit significant residuals (brighter regions) at structural boundaries. In sharp contrast, by decoupling phase and magnitude on the frequency manifold, EDNO yields the darkest and most uniform MAE maps, demonstrating superior spatial-spectral consistency with the ground truth. Furthermore, as shown in the full-resolution results (Fig.\ref{fig:Vis-WV2-full}), while methods like FusionMamba or MSCSCFormer possess much larger parameter counts, EDNO achieves highly competitive visual fidelity, effectively suppressing spectral bleeding. These observations confirm that our Euler-inspired interaction mechanism preserves natural color distributions while integrating sharp panchromatic textures more effectively than computationally heavier counterparts.

\begin{table}
\centering
\caption{Performance analysis of EDNO with varying iteration numbers $T$ of the Euler Feature Interaction Layer.}
\label{tab:config}
\centering
\setlength{\extrarowheight}{0pt}
\addtolength{\extrarowheight}{\aboverulesep}
\addtolength{\extrarowheight}{\belowrulesep}
\setlength{\aboverulesep}{0pt}
\setlength{\belowrulesep}{0pt}
\resizebox{\linewidth}{!}{
\begin{tabular}{c|ccccccccc}
\toprule
$T$   & PSNR$\uparrow$ & SSIM$\uparrow$ & $Q_4\uparrow$  & ERGAS$\downarrow$  & $D_s\downarrow$  \\ 
\midrule
2           & 41.1870 & 0.9724 & 0.8157  & 1.0925 &  0.0771  \\ 
3          & 41.4968 & 0.9738 & 0.8218  & 1.0472 &  0.0775  \\ 
\rowcolor[rgb]{0.922,0.922,0.922}
4          & \textbf{41.6488} & \textbf{0.9747} & \textbf{0.8262}  & \textbf{1.0259}  & \textbf{0.0768}  \\ 
5          & 41.4173 & 0.9733 & 0.8217  & 1.0527 &  0.0790  \\
\bottomrule
\end{tabular}
}
\label{tab:layer}
\end{table}

\begin{table}
\centering
\caption{Ablation study of different architectural configurations and internal components of EDNO.}
\label{tab:config}
\centering
\setlength{\extrarowheight}{0pt}
\addtolength{\extrarowheight}{\aboverulesep}
\addtolength{\extrarowheight}{\belowrulesep}
\setlength{\aboverulesep}{0pt}
\setlength{\belowrulesep}{0pt}
\resizebox{\linewidth}{!}{
\begin{tabular}{c|ccccccccc}
\toprule
Config   & PSNR$\uparrow$ & SSIM$\uparrow$ & $Q_4\uparrow$  & ERGAS$\downarrow$  & $D_s\downarrow$  \\ 
\midrule
(I)          & 38.7992  & 0.9552  & 0.7607   & 1.4415  & 0.0810  \\ 
(II)          & 38.9401  & 0.9551  & 0.7580   & 1.4080  & 0.0822  \\ 
(III)           & 37.4784 & 0.9434  & 0.7145  & 1.6593  & 0.0874  \\ 
(IV)           & 41.2360 & 0.9724 & 0.8150  & 1.0785  & 0.0789  \\ 
\rowcolor[rgb]{0.922,0.922,0.922}
\textbf{ours}          & \textbf{41.6488} & \textbf{0.9747} & \textbf{0.8262}  & \textbf{1.0259}  & \textbf{0.0768}  \\ 
\bottomrule
\end{tabular}
}
\label{tab:config}
\end{table}

\subsection{Ablation Study}
All ablation experiments are conducted on the WorldView-2 dataset to systematically evaluate the contribution of our EDNO.

\subsubsection{The Number of $T$}
Tab.\ref{tab:layer} illustrates the impact of the iteration number $T$ on the performance of EDNO. As $T$ increases from $2$ to $4$, the PSNR significantly improves from $41.1870$ dB to $41.6488$ dB. 
This trend suggests that progressive refinement in the frequency manifold yields superior pansharpening performance. However, further increasing the depth to $T=5$ results in a performance decline, with the spatial distortion index $D_s$ rising from $0.0768$ to $0.0790$. We attribute this to an over-smoothing effect where excessive iterations dilute fine-grained spectral features and introduce structural misalignments due to the high sensitivity of phase information. Consequently, EDNO employs $T=4$ to achieve an optimal balance between pansharpening fidelity and computational efficiency.

\subsubsection{Effectiveness of EFIL and its Components.}
Tab.\ref{tab:config} evaluates the contribution of different configurations to the overall performance. Replacing our EFIL with a vanilla Fourier Neural Operator (Config I) leads to a drastic PSNR drop from $41.6488$ dB to $38.7992$ dB, proving that direct complex weight multiplication is insufficient for the intricate spatial-spectral mapping required in pansharpening. The ablation of individual interaction modules further reveals their distinct roles, i.e., Config II (phase-only) and Config III (amplitude-only) both suffer from substantial performance degradation, confirming the indispensability of both phase and amplitude reconstructions for superior pansharpening.
Specifically, the severe drop in Config III highlights that omitting phase interaction leads to a loss of geometric fidelity. Furthermore, Config IV demonstrates that removing the $3 \times 3$ depth-wise convolution within the IFIM compromises the model's ability to capture local frequency correlations, resulting in inferior metrics. These results validate that the synergistic integration of EFIM and IFIM within the EFIL framework is essential for achieving high-fidelity pansharpening.

\begin{figure}[]
    \centering
    \includegraphics[width=\linewidth]{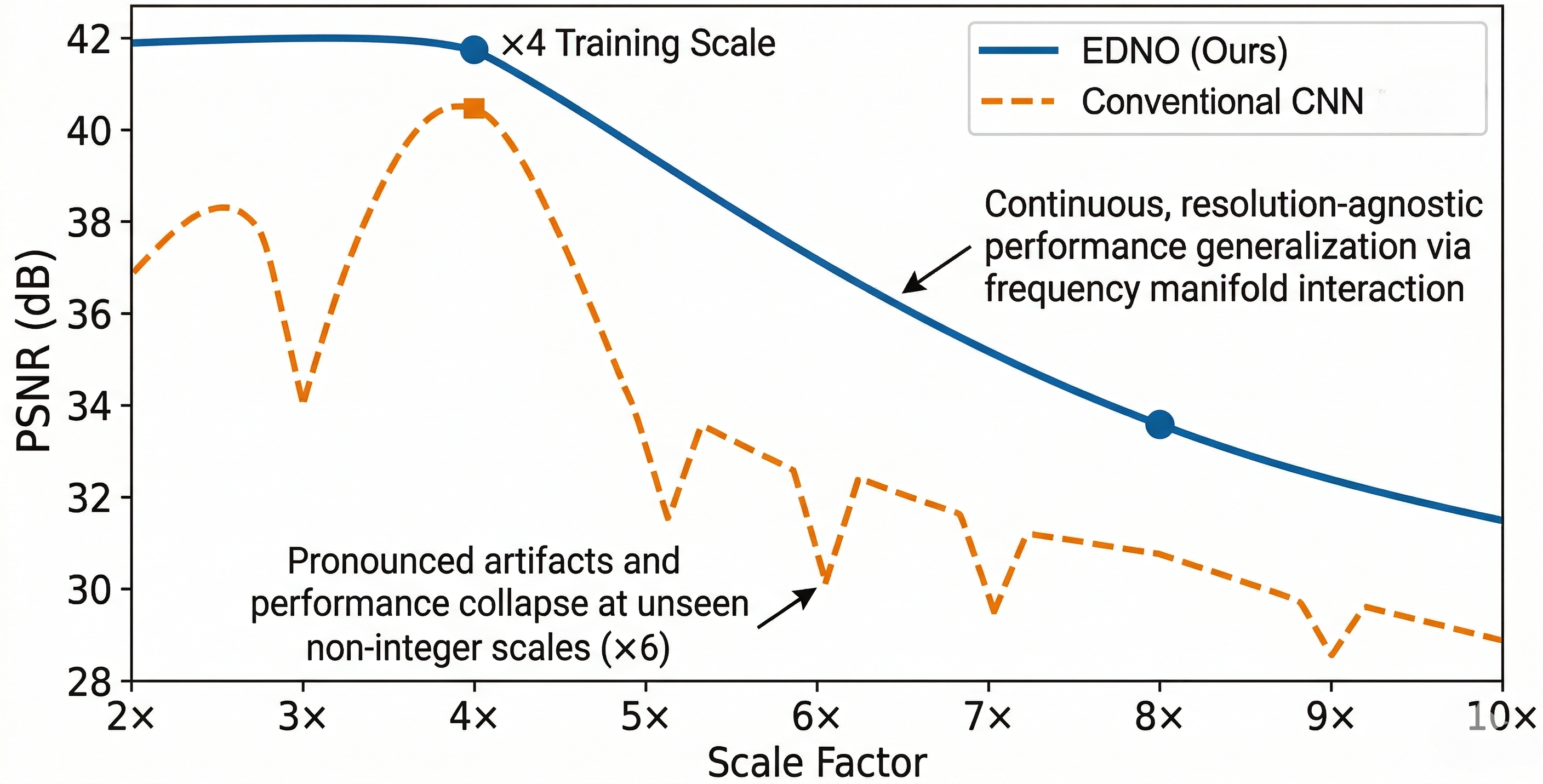}
    \caption{Zero-shot scale generalization from $\times 2$ to $\times 10$ resolution. }
    \label{fig:scale}
\end{figure}

\subsubsection{Continuous Scale Generalization.} To evaluate the resolution-agnostic nature of the proposed operator, we conduct a zero-shot generalization experiment across a resolution continuum from $\times 2$ to $\times 10$. As illustrated by the performance curves in Fig. 4, conventional discrete CNNs suffer from dramatic PSNR collapses at unseen non-integer scales (e.g., $\times 3, \times 6$) due to the spatial rigidity of fixed-grid convolutional kernels. In sharp contrast, EDNO maintains a monotonic and graceful performance decay, demonstrating seamless adaptability to arbitrary spatial sampling densities.

\subsubsection{Robustness to Resolution Perturbations.} 
\begin{figure}[]
    \centering
    \includegraphics[width=\linewidth]{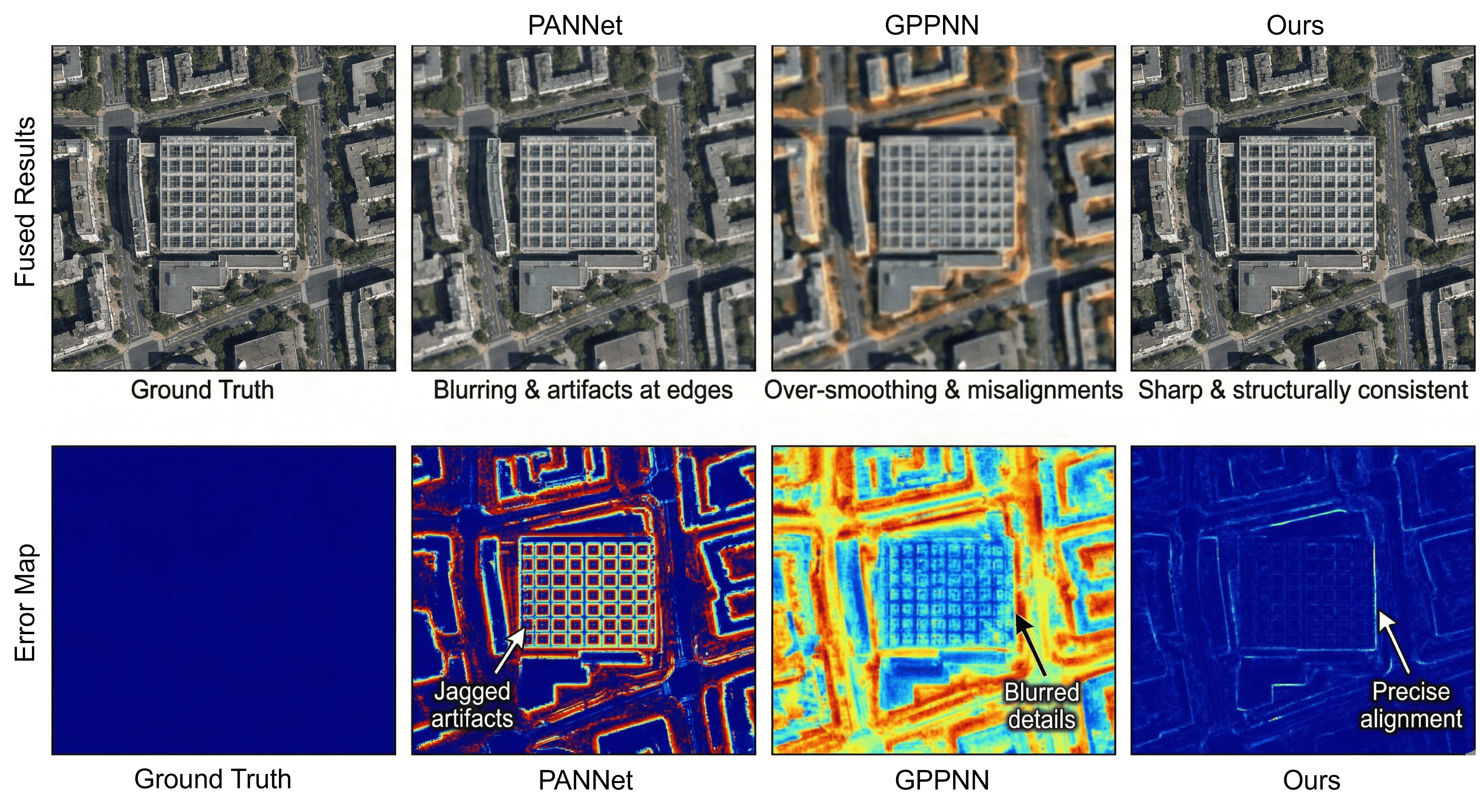}
    \caption{Qualitative comparison and absolute error maps under $5\%$ input scale perturbation. }
    \label{fig:drift}
\end{figure}

To evaluate EDNO under such geometric uncertainty, we introduce a $5\%$ scale jitter during inference and visualize the absolute error maps in Fig.\ref{fig:drift}. Conventional discrete CNNs and rigid physical priors exhibit significant performance collapse, manifesting as red high-error zones and jagged artifacts at structural boundaries due to the rigidity of convolutional kernels. In sharp contrast, EDNO maintains a remarkably clean error map with minimal residual magnitude. This robustness stems from the EFIM, which explicitly modulates the complex-valued frequency manifold to compensate for coordinate drifts.These results demonstrate that our Euler-inspired decoupling mechanism effectively captures the intrinsic geometric morphology rather than over-fitting to a fixed discrete grid.

\emph{For the more comprehensive analysis, additional experiments including cross-sensor ablation studies are provided in the Supplementary Material to further demonstrate the model's robustness.}

\section{Conclusion}

In this paper, we propose EDNO, a lightweight and principled framework for remote sensing pansharpening. By shifting the fusion task into a continuous frequency domain, we successfully decouple multimodal features into phase (geometric shapes) and magnitude (spectral intensity) via Euler’s formula. This mathematical simplification allows EDNO to resolve spatial-spectral conflicts through simple rotations and scaling. With only $246.9$ K parameters, EDNO achieves state-of-the-art performance and demonstrates a unique resolution-agnostic capability, enabling high-fidelity fusion across unseen scales.

%Moving forward, we aim to incorporate physics-based scattering models into the magnitude scaling branch to further enhance spectral consistency under extreme atmospheric conditions. Ultimately, shifting from black-box pixel mapping to principled mathematical interaction offers a promising path toward more interpretable and efficient Earth observation systems.

%While EDNO is theoretically continuous, our current training is conducted on fixed-scale datasets; exploring multi-scale or cross-sensor training could further unlock the zero-shot potential of our operator. Additionally, evaluating its robustness under extreme weather conditions (e.g., heavy cloud cover) remains a promising direction for future work.

\bibliographystyle{ACM-Reference-Format}
\bibliography{sample-sigconf-authordraft}


\begin{thebibliography}{60}

%%% the Web address.

\ifx \showCODEN    \undefined \def \showCODEN     #1{\unskip}     \fi
\ifx \showISBNx    \undefined \def \showISBNx     #1{\unskip}     \fi
\ifx \showISBNxiii \undefined \def \showISBNxiii  #1{\unskip}     \fi
\ifx \showISSN     \undefined \def \showISSN      #1{\unskip}     \fi
\ifx \showLCCN     \undefined \def \showLCCN      #1{\unskip}     \fi
\ifx \shownote     \undefined \def \shownote      #1{#1}          \fi
\ifx \showarticletitle \undefined \def \showarticletitle #1{#1}   \fi
\ifx \showURL      \undefined \def \showURL       {\relax}        \fi
% invisible to TeX
\providecommand\bibfield[2]{#2}
\providecommand\bibinfo[2]{#2}
\providecommand\natexlab[1]{#1}
\providecommand\showeprint[2][]{arXiv:#2}

\bibitem[Ahmad et~al\mbox{.}(2024)]%
        {ahmad2024fpga}
\bibfield{author}{\bibinfo{person}{Maruf Ahmad}, \bibinfo{person}{Lei Zhang}, {and} \bibinfo{person}{Muhammad~EH Chowdhury}.} \bibinfo{year}{2024}\natexlab{}.
\newblock \showarticletitle{Fpga implementation of complex-valued neural network for polar-represented image classification}.
\newblock \bibinfo{journal}{\emph{Sensors}} \bibinfo{volume}{24}, \bibinfo{number}{3} (\bibinfo{year}{2024}), \bibinfo{pages}{897}.
\newblock


\bibitem[Al-Qadasi and Waheed(2026)]%
        {al2026fourier}
\bibfield{author}{\bibinfo{person}{Basem Al-Qadasi} {and} \bibinfo{person}{Umair~Bin Waheed}.} \bibinfo{year}{2026}\natexlab{}.
\newblock \showarticletitle{Fourier neural operator for moonquake detection}.
\newblock \bibinfo{journal}{\emph{Earth and Space Science}} \bibinfo{volume}{13}, \bibinfo{number}{3} (\bibinfo{year}{2026}), \bibinfo{pages}{e2025EA004792}.
\newblock


\bibitem[Alparone et~al\mbox{.}(2008)]%
        {Alparone2008QNR}
\bibfield{author}{\bibinfo{person}{Luciano Alparone}, \bibinfo{person}{Bruno Aiazzi}, \bibinfo{person}{Stefano Baronti}, \bibinfo{person}{Andrea Garzelli}, \bibinfo{person}{Filippo Nencini}, {and} \bibinfo{person}{Massimo Selva}.} \bibinfo{year}{2008}\natexlab{}.
\newblock \showarticletitle{Multispectral and panchromatic data fusion assessment without reference}.
\newblock \bibinfo{journal}{\emph{Photogrammetric Engineering and Remote Sensing}}  \bibinfo{volume}{74} (\bibinfo{year}{2008}), \bibinfo{pages}{193--200}.
\newblock


\bibitem[Alparone et~al\mbox{.}(2007)]%
        {Alparone2007ERGAS}
\bibfield{author}{\bibinfo{person}{Luciano Alparone}, \bibinfo{person}{Lucien Wald}, \bibinfo{person}{Jocelyn Chanussot}, \bibinfo{person}{Claire Thomas}, \bibinfo{person}{Paolo Gamba}, {and} \bibinfo{person}{Lori~Mann Bruce}.} \bibinfo{year}{2007}\natexlab{}.
\newblock \showarticletitle{Comparison of Pansharpening Algorithms: Outcome of the 2006 GRS-S Data-Fusion Contest}.
\newblock \bibinfo{journal}{\emph{IEEE Transactions on Geoscience and Remote Sensing}} \bibinfo{volume}{45}, \bibinfo{number}{10} (\bibinfo{year}{2007}), \bibinfo{pages}{3012--3021}.
\newblock


\bibitem[Bandara and Patel(2022)]%
        {Bandara2022HyperTransformer}
\bibfield{author}{\bibinfo{person}{Wele Gedara~Chaminda Bandara} {and} \bibinfo{person}{Vishal~M. Patel}.} \bibinfo{year}{2022}\natexlab{}.
\newblock \showarticletitle{HyperTransformer: A Textural and Spectral Feature Fusion Transformer for Pansharpening}. In \bibinfo{booktitle}{\emph{2022 IEEE/CVF Conference on Computer Vision and Pattern Recognition (CVPR)}}. \bibinfo{pages}{1757--1767}.
\newblock


\bibitem[Cao et~al\mbox{.}(2024b)]%
        {cao2024zero}
\bibfield{author}{\bibinfo{person}{Qi Cao}, \bibinfo{person}{Liang-Jian Deng}, \bibinfo{person}{Wu Wang}, \bibinfo{person}{Junming Hou}, {and} \bibinfo{person}{Gemine Vivone}.} \bibinfo{year}{2024}\natexlab{b}.
\newblock \showarticletitle{Zero-shot semi-supervised learning for pansharpening}.
\newblock \bibinfo{journal}{\emph{Information Fusion}}  \bibinfo{volume}{101} (\bibinfo{year}{2024}), \bibinfo{pages}{102001}.
\newblock


\bibitem[Cao et~al\mbox{.}(2024a)]%
        {cao2024diffusion}
\bibfield{author}{\bibinfo{person}{Zihan Cao}, \bibinfo{person}{Shiqi Cao}, \bibinfo{person}{Liang-Jian Deng}, \bibinfo{person}{Xiao Wu}, \bibinfo{person}{Junming Hou}, {and} \bibinfo{person}{Gemine Vivone}.} \bibinfo{year}{2024}\natexlab{a}.
\newblock \showarticletitle{Diffusion model with disentangled modulations for sharpening multispectral and hyperspectral images}.
\newblock \bibinfo{journal}{\emph{Information Fusion}}  \bibinfo{volume}{104} (\bibinfo{year}{2024}), \bibinfo{pages}{102158}.
\newblock


\bibitem[Chen et~al\mbox{.}(2023)]%
        {Chen2023FFPN}
\bibfield{author}{\bibinfo{person}{Chaoyu Chen}, \bibinfo{person}{Xin Yang}, \bibinfo{person}{Rusi Chen}, \bibinfo{person}{Junxuan Yu}, \bibinfo{person}{Liwei Du}, \bibinfo{person}{Jian Wang}, \bibinfo{person}{Xindi Hu}, \bibinfo{person}{Yan Cao}, \bibinfo{person}{Yingying Liu}, {and} \bibinfo{person}{Dong Ni}.} \bibinfo{year}{2023}\natexlab{}.
\newblock \bibinfo{booktitle}{\emph{FFPN: Fourier Feature Pyramid Network for Ultrasound Image Segmentation}}.
\newblock \bibinfo{pages}{166--175}.
\newblock
\showISBNx{978-3-031-45672-5}


\bibitem[Choi et~al\mbox{.}(2010)]%
        {choi2010new}
\bibfield{author}{\bibinfo{person}{Jaewan Choi}, \bibinfo{person}{Kiyun Yu}, {and} \bibinfo{person}{Yongil Kim}.} \bibinfo{year}{2010}\natexlab{}.
\newblock \showarticletitle{A new adaptive component-substitution-based satellite image fusion by using partial replacement}.
\newblock \bibinfo{journal}{\emph{IEEE transactions on geoscience and remote sensing}} \bibinfo{volume}{49}, \bibinfo{number}{1} (\bibinfo{year}{2010}), \bibinfo{pages}{295--309}.
\newblock


\bibitem[Deng et~al\mbox{.}(2022)]%
        {deng2022machine}
\bibfield{author}{\bibinfo{person}{Liang-Jian Deng}, \bibinfo{person}{Gemine Vivone}, \bibinfo{person}{Mercedes~E Paoletti}, \bibinfo{person}{Giuseppe Scarpa}, \bibinfo{person}{Jiang He}, \bibinfo{person}{Yongjun Zhang}, \bibinfo{person}{Jocelyn Chanussot}, {and} \bibinfo{person}{Antonio Plaza}.} \bibinfo{year}{2022}\natexlab{}.
\newblock \showarticletitle{Machine learning in pansharpening: A benchmark, from shallow to deep networks}.
\newblock \bibinfo{journal}{\emph{IEEE Geoscience and Remote Sensing Magazine}} \bibinfo{volume}{10}, \bibinfo{number}{3} (\bibinfo{year}{2022}), \bibinfo{pages}{279--315}.
\newblock


\bibitem[Faragallah(2018)]%
        {Faragallah2018IHS}
\bibfield{author}{\bibinfo{person}{Osama~S. Faragallah}.} \bibinfo{year}{2018}\natexlab{}.
\newblock \showarticletitle{Enhancing multispectral imagery spatial resolution using optimized adaptive PCA and high-pass modulation}.
\newblock \bibinfo{journal}{\emph{International Journal of Remote Sensing}} \bibinfo{volume}{39}, \bibinfo{number}{20} (\bibinfo{year}{2018}), \bibinfo{pages}{6572--6586}.
\newblock
\href{https://doi.org/10.1080/01431161.2018.1463112}{doi:\nolinkurl{10.1080/01431161.2018.1463112}}


\bibitem[Garzelli and Nencini(2009)]%
        {Garzelli2009Q4}
\bibfield{author}{\bibinfo{person}{Andrea Garzelli} {and} \bibinfo{person}{Filippo Nencini}.} \bibinfo{year}{2009}\natexlab{}.
\newblock \showarticletitle{Hypercomplex Quality Assessment of Multi/Hyperspectral Images}.
\newblock \bibinfo{journal}{\emph{IEEE Geoscience and Remote Sensing Letters}} \bibinfo{volume}{6}, \bibinfo{number}{4} (\bibinfo{year}{2009}), \bibinfo{pages}{662--665}.
\newblock


\bibitem[Ghassemian(2016)]%
        {ghassemian2016review}
\bibfield{author}{\bibinfo{person}{Hassan Ghassemian}.} \bibinfo{year}{2016}\natexlab{}.
\newblock \showarticletitle{A review of remote sensing image fusion methods}.
\newblock \bibinfo{journal}{\emph{Information Fusion}}  \bibinfo{volume}{32} (\bibinfo{year}{2016}), \bibinfo{pages}{75--89}.
\newblock


\bibitem[Guibas et~al\mbox{.}(2021)]%
        {Guibas2021AFNO}
\bibfield{author}{\bibinfo{person}{John Guibas}, \bibinfo{person}{Morteza Mardani}, \bibinfo{person}{Zong-Yi Li}, \bibinfo{person}{Andrew Tao}, \bibinfo{person}{Anima Anandkumar}, {and} \bibinfo{person}{Bryan Catanzaro}.} \bibinfo{year}{2021}\natexlab{}.
\newblock \showarticletitle{Adaptive Fourier Neural Operators: Efficient Token Mixers for Transformers}.
\newblock \bibinfo{journal}{\emph{ArXiv}}  \bibinfo{volume}{abs/2111.13587} (\bibinfo{year}{2021}).
\newblock


\bibitem[He et~al\mbox{.}(2021)]%
        {he2021cnn}
\bibfield{author}{\bibinfo{person}{Lin He}, \bibinfo{person}{Jiawei Zhu}, \bibinfo{person}{Jun Li}, \bibinfo{person}{Antonio Plaza}, \bibinfo{person}{Jocelyn Chanussot}, {and} \bibinfo{person}{Zhuliang Yu}.} \bibinfo{year}{2021}\natexlab{}.
\newblock \showarticletitle{CNN-based hyperspectral pansharpening with arbitrary resolution}.
\newblock \bibinfo{journal}{\emph{IEEE Transactions on Geoscience and Remote Sensing}}  \bibinfo{volume}{60} (\bibinfo{year}{2021}), \bibinfo{pages}{1--21}.
\newblock


\bibitem[Horé and Ziou(2010)]%
        {Alain2010PSNR}
\bibfield{author}{\bibinfo{person}{Alain Horé} {and} \bibinfo{person}{Djemel Ziou}.} \bibinfo{year}{2010}\natexlab{}.
\newblock \showarticletitle{Image Quality Metrics: PSNR vs. SSIM}. In \bibinfo{booktitle}{\emph{2010 20th International Conference on Pattern Recognition}}. \bibinfo{pages}{2366--2369}.
\newblock


\bibitem[Hou et~al\mbox{.}(2023)]%
        {Hou2023BIM}
\bibfield{author}{\bibinfo{person}{Junming Hou}, \bibinfo{person}{Qi Cao}, \bibinfo{person}{Ran Ran}, \bibinfo{person}{Che Liu}, \bibinfo{person}{Junling Li}, {and} \bibinfo{person}{Liang-jian Deng}.} \bibinfo{year}{2023}\natexlab{}.
\newblock \showarticletitle{Bidomain Modeling Paradigm for Pansharpening}. In \bibinfo{booktitle}{\emph{Proceedings of the 31st ACM International Conference on Multimedia}} (Ottawa ON, Canada) \emph{(\bibinfo{series}{MM '23})}. \bibinfo{publisher}{Association for Computing Machinery}, \bibinfo{address}{New York, NY, USA}, \bibinfo{pages}{347–357}.
\newblock
\showISBNx{9798400701085}


\bibitem[Hou et~al\mbox{.}(2026)]%
        {hou2026nodiff}
\bibfield{author}{\bibinfo{person}{J. Hou}, \bibinfo{person}{R. Ran}, \bibinfo{person}{S. Chen}, \bibinfo{person}{Z. Chen}, \bibinfo{person}{X. Cong}, \bibinfo{person}{J. Li}, {and} \bibinfo{person}{L.-J. Deng}.} \bibinfo{year}{2026}\natexlab{}.
\newblock \showarticletitle{NODiff: Neural Operator Diffusion for Multispectral Image Fusion}.
\newblock \bibinfo{journal}{\emph{Proceedings of the AAAI Conference on Artificial Intelligence}} \bibinfo{volume}{40}, \bibinfo{number}{6} (\bibinfo{year}{2026}), \bibinfo{pages}{4753--4761}.
\newblock
\href{https://doi.org/10.1609/aaai.v40i6.42477}{doi:\nolinkurl{10.1609/aaai.v40i6.42477}}


\bibitem[Kashefi and Mukerji(2024)]%
        {kashefi2024novel}
\bibfield{author}{\bibinfo{person}{Ali Kashefi} {and} \bibinfo{person}{Tapan Mukerji}.} \bibinfo{year}{2024}\natexlab{}.
\newblock \showarticletitle{A novel Fourier neural operator framework for classification of multi-sized images: Application to three dimensional digital porous media}.
\newblock \bibinfo{journal}{\emph{Physics of Fluids}} \bibinfo{volume}{36}, \bibinfo{number}{5} (\bibinfo{year}{2024}).
\newblock


\bibitem[King and Wang(2001)]%
        {King2001wavelet}
\bibfield{author}{\bibinfo{person}{R.L. King} {and} \bibinfo{person}{Jianwen Wang}.} \bibinfo{year}{2001}\natexlab{}.
\newblock \showarticletitle{A wavelet based algorithm for pan sharpening Landsat 7 imagery}. In \bibinfo{booktitle}{\emph{IGARSS 2001. Scanning the Present and Resolving the Future. Proceedings. IEEE 2001 International Geoscience and Remote Sensing Symposium (Cat. No.01CH37217)}}, Vol.~\bibinfo{volume}{2}. \bibinfo{pages}{849--851 vol.2}.
\newblock


\bibitem[Li et~al\mbox{.}(2024)]%
        {li2024model}
\bibfield{author}{\bibinfo{person}{Jiaxin Li}, \bibinfo{person}{Ke Zheng}, \bibinfo{person}{Lianru Gao}, \bibinfo{person}{Li Ni}, \bibinfo{person}{Min Huang}, {and} \bibinfo{person}{Jocelyn Chanussot}.} \bibinfo{year}{2024}\natexlab{}.
\newblock \showarticletitle{Model-informed multistage unsupervised network for hyperspectral image super-resolution}.
\newblock \bibinfo{journal}{\emph{IEEE Transactions on Geoscience and Remote Sensing}}  \bibinfo{volume}{62} (\bibinfo{year}{2024}), \bibinfo{pages}{1--17}.
\newblock


\bibitem[Li et~al\mbox{.}(2023c)]%
        {li2023x}
\bibfield{author}{\bibinfo{person}{Jiaxin Li}, \bibinfo{person}{Ke Zheng}, \bibinfo{person}{Zhi Li}, \bibinfo{person}{Lianru Gao}, {and} \bibinfo{person}{Xiuping Jia}.} \bibinfo{year}{2023}\natexlab{c}.
\newblock \showarticletitle{X-shaped interactive autoencoders with cross-modality mutual learning for unsupervised hyperspectral image super-resolution}.
\newblock \bibinfo{journal}{\emph{IEEE transactions on geoscience and remote sensing}}  \bibinfo{volume}{61} (\bibinfo{year}{2023}), \bibinfo{pages}{1--17}.
\newblock


\bibitem[Li et~al\mbox{.}(2023a)]%
        {Li2023LGTEUN}
\bibfield{author}{\bibinfo{person}{Mingsong Li}, \bibinfo{person}{Yikun Liu}, \bibinfo{person}{Tao Xiao}, \bibinfo{person}{Yuwen Huang}, {and} \bibinfo{person}{Gongping Yang}.} \bibinfo{year}{2023}\natexlab{a}.
\newblock \showarticletitle{Local-Global Transformer Enhanced Unfolding Network for Pan-sharpening}. In \bibinfo{booktitle}{\emph{Proceedings of the Thirty-Second International Joint Conference on Artificial Intelligence, {IJCAI-23}}}, \bibfield{editor}{\bibinfo{person}{Edith Elkind}} (Ed.). \bibinfo{publisher}{International Joint Conferences on Artificial Intelligence Organization}, \bibinfo{pages}{1071--1079}.
\newblock
\newblock
\shownote{Main Track}.


\bibitem[Li et~al\mbox{.}(2023b)]%
        {li2023local}
\bibfield{author}{\bibinfo{person}{Mingsong Li}, \bibinfo{person}{Yikun Liu}, \bibinfo{person}{Tao Xiao}, \bibinfo{person}{Yuwen Huang}, {and} \bibinfo{person}{Gongping Yang}.} \bibinfo{year}{2023}\natexlab{b}.
\newblock \showarticletitle{Local-global transformer enhanced unfolding network for pan-sharpening}. In \bibinfo{booktitle}{\emph{Proceedings of the Thirty-Second International Joint Conference on Artificial Intelligence}}. \bibinfo{pages}{1071--1079}.
\newblock


\bibitem[Li et~al\mbox{.}(2020)]%
        {Li2020FNO}
\bibfield{author}{\bibinfo{person}{Zong-Yi Li}, \bibinfo{person}{Nikola~B. Kovachki}, \bibinfo{person}{Kamyar Azizzadenesheli}, \bibinfo{person}{Burigede Liu}, \bibinfo{person}{Kaushik Bhattacharya}, \bibinfo{person}{Andrew~M. Stuart}, {and} \bibinfo{person}{Anima Anandkumar}.} \bibinfo{year}{2020}\natexlab{}.
\newblock \showarticletitle{Fourier Neural Operator for Parametric Partial Differential Equations}.
\newblock \bibinfo{journal}{\emph{ArXiv}}  \bibinfo{volume}{abs/2010.08895} (\bibinfo{year}{2020}).
\newblock
\urldef\tempurl%
\url{https://api.semanticscholar.org/CorpusID:224705257}
\showURL{%
\tempurl}


\bibitem[Lin et~al\mbox{.}(2025)]%
        {lin2025alphapre}
\bibfield{author}{\bibinfo{person}{Kenghong Lin}, \bibinfo{person}{Baoquan Zhang}, \bibinfo{person}{Demin Yu}, \bibinfo{person}{Wenzhi Feng}, \bibinfo{person}{Shidong Chen}, \bibinfo{person}{Feifan Gao}, \bibinfo{person}{Xutao Li}, {and} \bibinfo{person}{Yunming Ye}.} \bibinfo{year}{2025}\natexlab{}.
\newblock \showarticletitle{AlphaPre: Amplitude-phase disentanglement model for precipitation nowcasting}. In \bibinfo{booktitle}{\emph{Proceedings of the Computer Vision and Pattern Recognition Conference}}. \bibinfo{pages}{17841--17850}.
\newblock


\bibitem[Liu(2000)]%
        {Liu2000SFIM}
\bibfield{author}{\bibinfo{person}{J.~G. Liu}.} \bibinfo{year}{2000}\natexlab{}.
\newblock \showarticletitle{Smoothing Filter-based Intensity Modulation: A spectral preserve image fusion technique for improving spatial details}.
\newblock \bibinfo{journal}{\emph{International Journal of Remote Sensing}} \bibinfo{volume}{21}, \bibinfo{number}{18} (\bibinfo{year}{2000}), \bibinfo{pages}{3461--3472}.
\newblock


\bibitem[Liu et~al\mbox{.}(2025)]%
        {liu2025PINO}
\bibfield{author}{\bibinfo{person}{Xinyang Liu}, \bibinfo{person}{Junming Hou}, \bibinfo{person}{Chenxu Wu}, \bibinfo{person}{Xiaofeng Cong}, \bibinfo{person}{Zihao Chen}, \bibinfo{person}{Shangqi Deng}, \bibinfo{person}{Junling Li}, \bibinfo{person}{Liang-Jian Deng}, {and} \bibinfo{person}{Bo Liu}.} \bibinfo{year}{2025}\natexlab{}.
\newblock \showarticletitle{Physics-informed Neural Operator for Pansharpening}. In \bibinfo{booktitle}{\emph{The Thirty-ninth Annual Conference on Neural Information Processing Systems}}.
\newblock
\urldef\tempurl%
\url{https://openreview.net/forum?id=tPI9Sw04sZ}
\showURL{%
\tempurl}


\bibitem[Liu et~al\mbox{.}(2020)]%
        {liu2020remote}
\bibfield{author}{\bibinfo{person}{Xiangyu Liu}, \bibinfo{person}{Qingjie Liu}, {and} \bibinfo{person}{Yunhong Wang}.} \bibinfo{year}{2020}\natexlab{}.
\newblock \showarticletitle{Remote sensing image fusion based on two-stream fusion network}.
\newblock \bibinfo{journal}{\emph{Information Fusion}}  \bibinfo{volume}{55} (\bibinfo{year}{2020}), \bibinfo{pages}{1--15}.
\newblock


\bibitem[Luo et~al\mbox{.}(2025)]%
        {luo2025pancomplex}
\bibfield{author}{\bibinfo{person}{Chunhui Luo}, \bibinfo{person}{Dong Li}, \bibinfo{person}{Xiaoliang Ma}, \bibinfo{person}{Xin Lu}, \bibinfo{person}{Zhiyuan Wang}, \bibinfo{person}{Jiangtong Tan}, {and} \bibinfo{person}{Xueyang Fu}.} \bibinfo{year}{2025}\natexlab{}.
\newblock \showarticletitle{PanComplex: leveraging complex-valued neural networks for enhanced pansharpening}. In \bibinfo{booktitle}{\emph{Proceedings of the Thirty-Fourth International Joint Conference on Artificial Intelligence}}. \bibinfo{pages}{1702--1710}.
\newblock


\bibitem[Masi et~al\mbox{.}(2016a)]%
        {masi2016pansharpening}
\bibfield{author}{\bibinfo{person}{Giuseppe Masi}, \bibinfo{person}{Davide Cozzolino}, \bibinfo{person}{Luisa Verdoliva}, {and} \bibinfo{person}{Giuseppe Scarpa}.} \bibinfo{year}{2016}\natexlab{a}.
\newblock \showarticletitle{Pansharpening by convolutional neural networks}.
\newblock \bibinfo{journal}{\emph{Remote Sensing}} \bibinfo{volume}{8}, \bibinfo{number}{7} (\bibinfo{year}{2016}), \bibinfo{pages}{594}.
\newblock


\bibitem[Masi et~al\mbox{.}(2016b)]%
        {Masi2016PNN}
\bibfield{author}{\bibinfo{person}{Giuseppe Masi}, \bibinfo{person}{Davide Cozzolino}, \bibinfo{person}{Luisa Verdoliva}, {and} \bibinfo{person}{Giuseppe Scarpa}.} \bibinfo{year}{2016}\natexlab{b}.
\newblock \showarticletitle{Pansharpening by Convolutional Neural Networks}.
\newblock \bibinfo{journal}{\emph{Remote Sensing}} \bibinfo{volume}{8}, \bibinfo{number}{7} (\bibinfo{year}{2016}).
\newblock
\showISSN{2072-4292}


\bibitem[Meng et~al\mbox{.}(2023)]%
        {meng2023pandiff}
\bibfield{author}{\bibinfo{person}{Qingyan Meng}, \bibinfo{person}{Wenxu Shi}, \bibinfo{person}{Sijia Li}, {and} \bibinfo{person}{Linlin Zhang}.} \bibinfo{year}{2023}\natexlab{}.
\newblock \showarticletitle{PanDiff: A novel pansharpening method based on denoising diffusion probabilistic model}.
\newblock \bibinfo{journal}{\emph{IEEE Transactions on Geoscience and Remote Sensing}}  \bibinfo{volume}{61} (\bibinfo{year}{2023}), \bibinfo{pages}{1--17}.
\newblock


\bibitem[Meng et~al\mbox{.}(2022)]%
        {meng2022vision}
\bibfield{author}{\bibinfo{person}{Xiangchao Meng}, \bibinfo{person}{Nan Wang}, \bibinfo{person}{Feng Shao}, {and} \bibinfo{person}{Shutao Li}.} \bibinfo{year}{2022}\natexlab{}.
\newblock \showarticletitle{Vision transformer for pansharpening}.
\newblock \bibinfo{journal}{\emph{IEEE Transactions on Geoscience and Remote Sensing}}  \bibinfo{volume}{60} (\bibinfo{year}{2022}), \bibinfo{pages}{1--11}.
\newblock


\bibitem[Sikdar et~al\mbox{.}(2022)]%
        {sikdar2022fully}
\bibfield{author}{\bibinfo{person}{Aniruddh Sikdar}, \bibinfo{person}{Sumanth Udupa}, \bibinfo{person}{Suresh Sundaram}, {and} \bibinfo{person}{Narasimhan Sundararajan}.} \bibinfo{year}{2022}\natexlab{}.
\newblock \showarticletitle{Fully complex-valued fully convolutional multi-feature fusion network (fc 2 mfn) for building segmentation of insar images}. In \bibinfo{booktitle}{\emph{2022 IEEE Symposium Series on Computational Intelligence (SSCI)}}. IEEE, \bibinfo{pages}{581--587}.
\newblock


\bibitem[Thomas et~al\mbox{.}(2008)]%
        {thomas2008synthesis}
\bibfield{author}{\bibinfo{person}{Claire Thomas}, \bibinfo{person}{Thierry Ranchin}, \bibinfo{person}{Lucien Wald}, {and} \bibinfo{person}{Jocelyn Chanussot}.} \bibinfo{year}{2008}\natexlab{}.
\newblock \showarticletitle{Synthesis of multispectral images to high spatial resolution: A critical review of fusion methods based on remote sensing physics}.
\newblock \bibinfo{journal}{\emph{IEEE Transactions on Geoscience and Remote Sensing}} \bibinfo{volume}{46}, \bibinfo{number}{5} (\bibinfo{year}{2008}), \bibinfo{pages}{1301--1312}.
\newblock


\bibitem[Tian et~al\mbox{.}(2023)]%
        {Tian2023EulerNet}
\bibfield{author}{\bibinfo{person}{Zhen Tian}, \bibinfo{person}{Ting Bai}, \bibinfo{person}{Wayne~Xin Zhao}, \bibinfo{person}{Ji-Rong Wen}, {and} \bibinfo{person}{Zhao Cao}.} \bibinfo{year}{2023}\natexlab{}.
\newblock \showarticletitle{EulerNet: Adaptive Feature Interaction Learning via Euler's Formula for CTR Prediction} \emph{(\bibinfo{series}{SIGIR '23})}. \bibinfo{publisher}{Association for Computing Machinery}, \bibinfo{address}{New York, NY, USA}, \bibinfo{pages}{1376–1385}.
\newblock
\showISBNx{9781450394086}
\href{https://doi.org/10.1145/3539618.3591681}{doi:\nolinkurl{10.1145/3539618.3591681}}


\bibitem[Vivone et~al\mbox{.}(2024)]%
        {vivone2024deep}
\bibfield{author}{\bibinfo{person}{Gemine Vivone}, \bibinfo{person}{Liang-Jian Deng}, \bibinfo{person}{Shangqi Deng}, \bibinfo{person}{Danfeng Hong}, \bibinfo{person}{Menghui Jiang}, \bibinfo{person}{Chenyu Li}, \bibinfo{person}{Wei Li}, \bibinfo{person}{Huanfeng Shen}, \bibinfo{person}{Xiao Wu}, \bibinfo{person}{Jin-Liang Xiao}, {et~al\mbox{.}}} \bibinfo{year}{2024}\natexlab{}.
\newblock \showarticletitle{Deep learning in remote sensing image fusion: Methods, protocols, data, and future perspectives}.
\newblock \bibinfo{journal}{\emph{IEEE geoscience and remote sensing magazine}} \bibinfo{volume}{13}, \bibinfo{number}{1} (\bibinfo{year}{2024}), \bibinfo{pages}{269--310}.
\newblock


\bibitem[Wald et~al\mbox{.}(1997)]%
        {wald1997fusion}
\bibfield{author}{\bibinfo{person}{Lucien Wald}, \bibinfo{person}{Thierry Ranchin}, {and} \bibinfo{person}{Marc Mangolini}.} \bibinfo{year}{1997}\natexlab{}.
\newblock \showarticletitle{Fusion of satellite images of different spatial resolutions: Assessing the quality of resulting images}.
\newblock \bibinfo{journal}{\emph{Photogrammetric engineering and remote sensing}} \bibinfo{volume}{63}, \bibinfo{number}{6} (\bibinfo{year}{1997}), \bibinfo{pages}{691--699}.
\newblock


\bibitem[Wang et~al\mbox{.}(2024b)]%
        {wang2024zero}
\bibfield{author}{\bibinfo{person}{Hebaixu Wang}, \bibinfo{person}{Hao Zhang}, \bibinfo{person}{Xin Tian}, {and} \bibinfo{person}{Jiayi Ma}.} \bibinfo{year}{2024}\natexlab{b}.
\newblock \showarticletitle{Zero-sharpen: A universal pansharpening method across satellites for reducing scale-variance gap via zero-shot variation}.
\newblock \bibinfo{journal}{\emph{Information Fusion}}  \bibinfo{volume}{101} (\bibinfo{year}{2024}), \bibinfo{pages}{102003}.
\newblock


\bibitem[Wang et~al\mbox{.}(2024a)]%
        {wang2024general}
\bibfield{author}{\bibinfo{person}{Wu Wang}, \bibinfo{person}{Liang-Jian Deng}, \bibinfo{person}{Ran Ran}, {and} \bibinfo{person}{Gemine Vivone}.} \bibinfo{year}{2024}\natexlab{a}.
\newblock \showarticletitle{A general paradigm with detail-preserving conditional invertible network for image fusion}.
\newblock \bibinfo{journal}{\emph{International Journal of Computer Vision}} \bibinfo{volume}{132}, \bibinfo{number}{4} (\bibinfo{year}{2024}), \bibinfo{pages}{1029--1054}.
\newblock


\bibitem[Wang et~al\mbox{.}(2004)]%
        {Zhou2004SSIM}
\bibfield{author}{\bibinfo{person}{Zhou Wang}, \bibinfo{person}{A.C. Bovik}, \bibinfo{person}{H.R. Sheikh}, {and} \bibinfo{person}{E.P. Simoncelli}.} \bibinfo{year}{2004}\natexlab{}.
\newblock \showarticletitle{Image quality assessment: from error visibility to structural similarity}.
\newblock \bibinfo{journal}{\emph{IEEE Transactions on Image Processing}} \bibinfo{volume}{13}, \bibinfo{number}{4} (\bibinfo{year}{2004}), \bibinfo{pages}{600--612}.
\newblock


\bibitem[Wei and Zhang(2023)]%
        {wei2023super}
\bibfield{author}{\bibinfo{person}{Min Wei} {and} \bibinfo{person}{Xuesong Zhang}.} \bibinfo{year}{2023}\natexlab{}.
\newblock \showarticletitle{Super-resolution neural operator}. In \bibinfo{booktitle}{\emph{Proceedings of the IEEE/CVF Conference on Computer Vision and Pattern Recognition}}. \bibinfo{pages}{18247--18256}.
\newblock


\bibitem[Wu et~al\mbox{.}(2021)]%
        {wu2021dynamic}
\bibfield{author}{\bibinfo{person}{Xiao Wu}, \bibinfo{person}{Ting-Zhu Huang}, \bibinfo{person}{Liang-Jian Deng}, {and} \bibinfo{person}{Tian-Jing Zhang}.} \bibinfo{year}{2021}\natexlab{}.
\newblock \showarticletitle{Dynamic cross feature fusion for remote sensing pansharpening}. In \bibinfo{booktitle}{\emph{Proceedings of the IEEE/CVF International Conference on Computer Vision}}. \bibinfo{pages}{14687--14696}.
\newblock


\bibitem[Xia et~al\mbox{.}(2026)]%
        {xia2026swift}
\bibfield{author}{\bibinfo{person}{Zeyu Xia}, \bibinfo{person}{Chenxi Sun}, \bibinfo{person}{Tianyu Xin}, \bibinfo{person}{Yubo Zeng}, \bibinfo{person}{Haoyu Chen}, {and} \bibinfo{person}{Liang-Jian Deng}.} \bibinfo{year}{2026}\natexlab{}.
\newblock \showarticletitle{Swift: A general sensitive weight identification framework for fast sensor-transfer pansharpening}. In \bibinfo{booktitle}{\emph{Proceedings of the AAAI Conference on Artificial Intelligence}}, Vol.~\bibinfo{volume}{40}. \bibinfo{pages}{10960--10968}.
\newblock


\bibitem[Xie et~al\mbox{.}(2024)]%
        {Xie2024FusionMamba}
\bibfield{author}{\bibinfo{person}{Xinyu Xie}, \bibinfo{person}{Yawen Cui}, \bibinfo{person}{Tao Tan}, \bibinfo{person}{Xubin Zheng}, {and} \bibinfo{person}{Zitong Yu}.} \bibinfo{year}{2024}\natexlab{}.
\newblock \showarticletitle{FusionMamba: dynamic feature enhancement for multimodal image fusion with Mamba}.
\newblock \bibinfo{journal}{\emph{Visual Intelligence}}  \bibinfo{volume}{2} (\bibinfo{year}{2024}).
\newblock
\urldef\tempurl%
\url{https://api.semanticscholar.org/CorpusID:269148750}
\showURL{%
\tempurl}


\bibitem[Xu et~al\mbox{.}(2021)]%
        {Xu2021GPPNN}
\bibfield{author}{\bibinfo{person}{Shuang Xu}, \bibinfo{person}{Jiangshe Zhang}, \bibinfo{person}{Zixiang Zhao}, \bibinfo{person}{Kai Sun}, \bibinfo{person}{Junmin Liu}, {and} \bibinfo{person}{Chunxia Zhang}.} \bibinfo{year}{2021}\natexlab{}.
\newblock \showarticletitle{Deep Gradient Projection Networks for Pan-sharpening}. In \bibinfo{booktitle}{\emph{2021 IEEE/CVF Conference on Computer Vision and Pattern Recognition (CVPR)}}. \bibinfo{pages}{1366--1375}.
\newblock


\bibitem[Yang et~al\mbox{.}(2017)]%
        {Yang2017PanNet}
\bibfield{author}{\bibinfo{person}{Junfeng Yang}, \bibinfo{person}{Xueyang Fu}, \bibinfo{person}{Yuwen Hu}, \bibinfo{person}{Yue Huang}, \bibinfo{person}{Xinghao Ding}, {and} \bibinfo{person}{John Paisley}.} \bibinfo{year}{2017}\natexlab{}.
\newblock \showarticletitle{PanNet: A Deep Network Architecture for Pan-Sharpening}. In \bibinfo{booktitle}{\emph{2017 IEEE International Conference on Computer Vision (ICCV)}}. \bibinfo{pages}{1753--1761}.
\newblock


\bibitem[Ye et~al\mbox{.}(2024)]%
        {Ye2024MSCSCformer}
\bibfield{author}{\bibinfo{person}{Yongxu Ye}, \bibinfo{person}{Tingting Wang}, \bibinfo{person}{Faming Fang}, {and} \bibinfo{person}{Guixu Zhang}.} \bibinfo{year}{2024}\natexlab{}.
\newblock \showarticletitle{MSCSCformer: Multiscale Convolutional Sparse Coding-Based Transformer for Pansharpening}.
\newblock \bibinfo{journal}{\emph{IEEE Transactions on Geoscience and Remote Sensing}}  \bibinfo{volume}{62} (\bibinfo{year}{2024}), \bibinfo{pages}{1--12}.
\newblock


\bibitem[Yin et~al\mbox{.}(2025)]%
        {yin2025cascaded}
\bibfield{author}{\bibinfo{person}{Junru Yin}, \bibinfo{person}{Zhiheng Huang}, \bibinfo{person}{Qiqiang Chen}, \bibinfo{person}{Wei Huang}, \bibinfo{person}{Le Sun}, \bibinfo{person}{Qinggang Wu}, {and} \bibinfo{person}{Ruixia Hou}.} \bibinfo{year}{2025}\natexlab{}.
\newblock \showarticletitle{Cascaded Local--Nonlocal Pansharpening with Adaptive Channel-Kernel Convolution and Multi-Scale Large-Kernel Attention}.
\newblock \bibinfo{journal}{\emph{Remote Sensing}} \bibinfo{volume}{18}, \bibinfo{number}{1} (\bibinfo{year}{2025}), \bibinfo{pages}{97}.
\newblock


\bibitem[Yuhas et~al\mbox{.}(1992)]%
        {Yuhas1992SAM}
\bibfield{author}{\bibinfo{person}{Roberta~H. Yuhas}, \bibinfo{person}{Alexander F.~H. Goetz}, {and} \bibinfo{person}{Joseph~W. Boardman}.} \bibinfo{year}{1992}\natexlab{}.
\newblock \showarticletitle{Discrimination among semi-arid landscape endmembers using the Spectral Angle Mapper (SAM) algorithm}. In \bibinfo{booktitle}{\emph{JPL, Summaries of the Third Annual JPL Airborne Geoscience Workshop. Volume 1: AVIRIS Workshop}}. \bibinfo{pages}{147--149}.
\newblock


\bibitem[Zeng et~al\mbox{.}(2025)]%
        {Zeng2025Cross-Modal}
\bibfield{author}{\bibinfo{person}{Haoying Zeng}, \bibinfo{person}{Xiaoyuan Yang}, \bibinfo{person}{Kangqing Shen}, \bibinfo{person}{Yixiao Li}, \bibinfo{person}{Jin Jiang}, {and} \bibinfo{person}{Fangyi Li}.} \bibinfo{year}{2025}\natexlab{}.
\newblock \showarticletitle{Cross-Modal Contrastive Pansharpening via Uncertainty Guidance}.
\newblock \bibinfo{journal}{\emph{IEEE Transactions on Geoscience and Remote Sensing}}  \bibinfo{volume}{63} (\bibinfo{year}{2025}), \bibinfo{pages}{1--14}.
\newblock


\bibitem[Zhang et~al\mbox{.}(2022)]%
        {zhang2022progress}
\bibfield{author}{\bibinfo{person}{Bing Zhang}, \bibinfo{person}{Yuanfeng Wu}, \bibinfo{person}{Boya Zhao}, \bibinfo{person}{Jocelyn Chanussot}, \bibinfo{person}{Danfeng Hong}, \bibinfo{person}{Jing Yao}, {and} \bibinfo{person}{Lianru Gao}.} \bibinfo{year}{2022}\natexlab{}.
\newblock \showarticletitle{Progress and challenges in intelligent remote sensing satellite systems}.
\newblock \bibinfo{journal}{\emph{IEEE Journal of Selected Topics in Applied Earth Observations and Remote Sensing}}  \bibinfo{volume}{15} (\bibinfo{year}{2022}), \bibinfo{pages}{1814--1822}.
\newblock


\bibitem[Zhang et~al\mbox{.}(2025b)]%
        {zhang2025s2wmamba}
\bibfield{author}{\bibinfo{person}{Haoyu Zhang}, \bibinfo{person}{Junhan Luo}, \bibinfo{person}{Yugang Cao}, \bibinfo{person}{Jie Huang}, {et~al\mbox{.}}} \bibinfo{year}{2025}\natexlab{b}.
\newblock \showarticletitle{S2WMamba: A Spectral-Spatial Wavelet Mamba for Pansharpening}.
\newblock \bibinfo{journal}{\emph{arXiv preprint arXiv:2512.06330}} (\bibinfo{year}{2025}).
\newblock


\bibitem[Zhang et~al\mbox{.}(2025a)]%
        {zhang2025rethinking}
\bibfield{author}{\bibinfo{person}{Ran Zhang}, \bibinfo{person}{Xuanhua He}, \bibinfo{person}{Li Xueheng}, \bibinfo{person}{Ke Cao}, \bibinfo{person}{Liu Liu}, \bibinfo{person}{Wenbo Xu}, \bibinfo{person}{Fang Jiabin}, \bibinfo{person}{Yang Qize}, {and} \bibinfo{person}{Jie Zhang}.} \bibinfo{year}{2025}\natexlab{a}.
\newblock \showarticletitle{Rethinking Pan-sharpening: Principled Design, Unified Training, and a Universal Loss Surpass Brute-Force Scaling}.
\newblock \bibinfo{journal}{\emph{arXiv e-prints}} (\bibinfo{year}{2025}), \bibinfo{pages}{arXiv--2507}.
\newblock


\bibitem[Zhang et~al\mbox{.}(2024)]%
        {zhang2024dmfourllie}
\bibfield{author}{\bibinfo{person}{Tongshun Zhang}, \bibinfo{person}{Pingping Liu}, \bibinfo{person}{Ming Zhao}, {and} \bibinfo{person}{Haotian Lv}.} \bibinfo{year}{2024}\natexlab{}.
\newblock \showarticletitle{Dmfourllie: Dual-stage and multi-branch fourier network for low-light image enhancement}. In \bibinfo{booktitle}{\emph{Proceedings of the 32nd ACM international conference on multimedia}}. \bibinfo{pages}{7434--7443}.
\newblock


\bibitem[Zheng et~al\mbox{.}(2023)]%
        {zheng2023deep}
\bibfield{author}{\bibinfo{person}{Kaiwen Zheng}, \bibinfo{person}{Jie Huang}, \bibinfo{person}{Man Zhou}, \bibinfo{person}{Danfeng Hong}, {and} \bibinfo{person}{Feng Zhao}.} \bibinfo{year}{2023}\natexlab{}.
\newblock \showarticletitle{Deep adaptive pansharpening via uncertainty-aware image fusion}.
\newblock \bibinfo{journal}{\emph{IEEE Transactions on Geoscience and Remote Sensing}}  \bibinfo{volume}{61} (\bibinfo{year}{2023}), \bibinfo{pages}{1--15}.
\newblock


\bibitem[Zhou et~al\mbox{.}(2022c)]%
        {Zhou2022PanFormer}
\bibfield{author}{\bibinfo{person}{Huanyu Zhou}, \bibinfo{person}{Qingjie Liu}, {and} \bibinfo{person}{Yunhong Wang}.} \bibinfo{year}{2022}\natexlab{c}.
\newblock \showarticletitle{PanFormer: A Transformer Based Model for Pan-Sharpening}. In \bibinfo{booktitle}{\emph{2022 IEEE International Conference on Multimedia and Expo (ICME)}}. \bibinfo{pages}{1--6}.
\newblock


\bibitem[Zhou et~al\mbox{.}(2022a)]%
        {Zhou2022CTINN}
\bibfield{author}{\bibinfo{person}{Man Zhou}, \bibinfo{person}{Jie Huang}, \bibinfo{person}{Yanchi Fang}, \bibinfo{person}{Xueyang Fu}, {and} \bibinfo{person}{Aiping Liu}.} \bibinfo{year}{2022}\natexlab{a}.
\newblock \showarticletitle{Pan-Sharpening with Customized Transformer and Invertible Neural Network}.
\newblock \bibinfo{journal}{\emph{Proceedings of the AAAI Conference on Artificial Intelligence}}  \bibinfo{volume}{36} (\bibinfo{date}{06} \bibinfo{year}{2022}), \bibinfo{pages}{3553--3561}.
\newblock


\bibitem[Zhou et~al\mbox{.}(2022b)]%
        {Zhou2022SFIIN}
\bibfield{author}{\bibinfo{person}{Man Zhou}, \bibinfo{person}{Jie Huang}, \bibinfo{person}{Keyu Yan}, \bibinfo{person}{Hu Yu}, \bibinfo{person}{Xueyang Fu}, \bibinfo{person}{Aiping Liu}, \bibinfo{person}{Xian Wei}, {and} \bibinfo{person}{Feng Zhao}.} \bibinfo{year}{2022}\natexlab{b}.
\newblock \showarticletitle{Spatial-Frequency Domain Information Integration for Pan-Sharpening}. In \bibinfo{booktitle}{\emph{Computer Vision - {ECCV} 2022 - 17th European Conference, Tel Aviv, Israel, October 23-27, 2022, Proceedings, Part {XVIII}}} \emph{(\bibinfo{series}{Lecture Notes in Computer Science})}. \bibinfo{publisher}{Springer}, \bibinfo{pages}{274--291}.
\newblock


\end{thebibliography}

\end{document}